\documentclass[11pt]{article}

\PassOptionsToPackage{table}{xcolor}

\usepackage[final]{acl}
\providecommand{\linenumbers}{}
\providecommand{\nolinenumbers}{}

\makeatletter
\let\old@makecol\@makecol
\renewcommand{\@makecol}{%
   \let\oldmc@result\mc@result
   \let\mc@result\@empty
   \old@makecol
   \let\mc@result\oldmc@result
}
\makeatother
\usepackage{etoolbox}
\AtBeginEnvironment{figure}{\nolinenumbers}
\AtBeginEnvironment{figure*}{\nolinenumbers}
\AtBeginEnvironment{table}{\nolinenumbers}
\AtBeginEnvironment{table*}{\nolinenumbers}
\AtBeginEnvironment{tcolorbox}{\nolinenumbers}
\AfterEndEnvironment{tcolorbox}{\linenumbers}

\usepackage{times}
\usepackage{latexsym}
\usepackage{amsmath}
\usepackage{amssymb}
\usepackage{booktabs}
\usepackage{multirow}
\usepackage{graphicx}
\usepackage{enumitem}
\usepackage{tcolorbox}
\tcbset{before={\par\nolinenumbers\noindent}, after={\par\linenumbers}, width=\columnwidth}

\definecolor{mycolor}{RGB}{220,235,255}
\definecolor{findingbg}{RGB}{245,245,245}
\usepackage[T1]{fontenc}
\usepackage[utf8]{inputenc}

\usepackage{microtype}

\usepackage{graphicx}

\newcommand{\ours}{\textsc{Prefilling-dLLM}}

\title{Prefilling-dLLM: Predictive Prefilling for Long-Context Inference in Diffusion Language Models}

\author{
  \textbf{Jing Xiong}\textsuperscript{1} \quad
  \textbf{Qi Han}\textsuperscript{1} \quad
  \textbf{Shansan Gong}\textsuperscript{1} \quad
  \textbf{Yunta Hsieh}\textsuperscript{2} \\
  \textbf{Boyuan Zheng}\textsuperscript{2} \quad
  \textbf{Chengyue Wu}\textsuperscript{1} \quad
  \textbf{Chaofan Tao}\textsuperscript{1} \quad
  \textbf{Chenyang Zhao}\textsuperscript{3} \quad
  \textbf{Ngai Wong}\textsuperscript{1} \\
  \textsuperscript{1}The University of Hong Kong \quad
  \textsuperscript{2}University of Michigan, Ann Arbor \quad
  \textsuperscript{3}LMSYS Org
}

\begin{document}
\maketitle
\begin{abstract}
Diffusion large language models (dLLMs) re-encode the entire prefix at every denoising step, causing recomputation that scales quadratically with context length and becomes prohibitive for long-context scenarios. We propose \ours{}, a \emph{training-free} prefill-decode disaggregation framework for dLLMs that partitions the prefix into $N$ chunks, caches their KV representations once, and selects the top-$K$ chunks with intra-chunk token sparsity for decoding, showing that sparse attention can outperform dense attention while reducing complexity from $O((L_p + L_q + L_d)^2 \cdot T)$ to $O\bigl(N(C+L_q+|\mathbf{m}|)^2 + (L_d^2 + KC)T\bigr)$. On LongBench and InfiniteBench, \ours{} achieves state-of-the-art quality among dLLM acceleration methods with 9.1--28.0$\times$ speedup at 8K--32K contexts, and we identify the memory bottleneck introduced by caching prefix KV across denoising steps. We further show that beginning-of-sequence tokens prepended to each chunk act as periodic attention anchors that eliminate the lost-in-the-middle phenomenon.\footnote{Our code is available at \url{https://github.com/menik1126/Prefilling-dLLM}.}
\end{abstract}

\section{Introduction}
\vspace{-2mm}
Diffusion large language models (dLLMs) have emerged as a promising alternative to autoregressive (AR) models, offering the ability to generate multiple tokens in parallel through iterative denoising~\citep{nie2025llada, ye2025dream7bdiffusionlarge, sahoo2024simple, austin2021structured}. Unlike AR models that produce tokens sequentially from left to right, dLLMs corrupt and reconstruct entire sequences simultaneously, enabling flexible generation orders and potentially faster inference~\citep{wu2025fastdllmtrainingfreeaccelerationdiffusion, wang2025diffusionllmsfasterthanarinference}. However, this paradigm introduces a critical inefficiency in long-context scenarios: the entire input prefix must participate in every denoising step, even though its representation remains largely unchanged across iterations.

In autoregressive LLM serving, the \emph{prefill-decode disaggregation} architecture~\citep{zhong2024distserve} assigns the prefill and decode phases to separate GPU clusters, exploiting their distinct computational profiles (prefill is compute-bound while decode is memory-bound) to maximize hardware utilization and serving throughput. In contrast, dLLM inference is fundamentally compute-bound throughout: since the entire sequence (prefix + decode) must be jointly processed at every denoising step, each iteration performs a full forward pass over the combined sequence, making the workload dominated by matrix multiplications rather than \emph{memory bandwidth}. This compute-bound nature persists across all denoising iterations, unlike AR decoding where only a single new token is appended per step. Recent work on dLLM acceleration has explored KV caching strategies~\citep{ma2026dkv, liu2025dllmcacheacceleratingdiffusionlarge, nguyentri2025attentionneedkvcache} and sparse attention mechanisms~\citep{wang2025sparsed, song2025sparsedllmacceleratingdiffusionllms, jiang2025d2cacheacceleratingdiffusionbasedllms}, yet none explores disaggregating the prefill and decode stages to avoid repeated long-context computation across denoising iterations.

Our key insight is that in long-context dLLM inference, the input prefix is redundantly processed at every denoising iteration, yet attention from response tokens to the prefix exhibits strong locality bias that intensifies across steps, and only a small fraction of prefix tokens are actively attended to. Motivated by this observation, we present \ours{} (\textbf{Prefilling} for \textbf{d}iffusion \textbf{LLM}s), which computes the prefix KV cache once in a dedicated prefill stage and reuses it across all decode steps without recomputation. Specifically, we partition the prefix into $N$ fixed-size chunks of size $C$ with intra-chunk attention, reducing prefill complexity from $O(L_p^2)$ to $O(N \cdot C^2)$ and enabling parallel processing across devices. During decode, we retain only a small subset of relevant chunks via retrieval-augmented generation~\citep{jiang2024minference, lai2025flexprefill, xu2025xattention, yuan2025nativesparseattentionhardwarealigned}, reducing complexity from $O((L_p + L_q + L_d)^2 \cdot T)$ to $O\bigl(N(C+L_q+|\mathbf{m}|)^2 + (L_d^2 + KC)T\bigr)$ while preserving quality.

We evaluate \ours{} on several long-context benchmarks~\citep{bai2024longbench, zhang2024bench, kamradt2023niah} and demonstrate that our approach achieves substantial speedups over naive dLLM inference while maintaining generation quality. Our contributions are as follows:
\vspace{-2mm}
\begin{itemize}[leftmargin=*]
    \item We propose \ours{}, a \emph{training-free prefill-decode disaggregation} framework for dLLMs. By prefilling the prefix KV cache once and sharing it across all denoising iterations, we eliminate recomputation and achieve significant speedups that scale with context length.
    \item We introduce \emph{sparse prefilling} that selects relevant chunks and tokens, reducing complexity to $O\bigl(N(C+L_q+|\mathbf{m}|)^2 + (L_d^2 + KC)T\bigr)$ with up to 28$\times$ end-to-end speedup at 32K.
    \item We show that beginning-of-sequence (BOS) tokens prepended to each chunk act as periodic attention anchors, mitigating the lost-in-the-middle phenomenon in dLLMs without introducing attention sinks.
\end{itemize}
\vspace{-2mm}

\vspace{-2mm}
\section{Related Work}
\vspace{-2mm}
\subsection{Diffusion Language Models}

Diffusion models have been extended from continuous domains to discrete text generation through various formulations. Early work explored continuous diffusion over word embeddings~\citep{li2022diffusion, gong2022diffuseq} and masked diffusion over discrete tokens~\citep{austin2021structured, he2023diffusionbert, sahoo2024simple}. More recently, masked discrete diffusion has been scaled to large language models~\citep{gong2025scalingdiffusionlanguagemodels}: LLaDA~\citep{nie2025llada} demonstrated that masked diffusion can match autoregressive models at the 8B parameter scale, while Dream~\citep{ye2025dream7bdiffusionlarge} and MDLM~\citep{sahoo2024simple} further validated the effectiveness of this paradigm. Subsequent efforts have focused on scaling~\citep{bie2025llada20scalingdiffusionlanguage, gong2025scalingdiffusionlanguagemodels}, preference alignment~\citep{zhu2025llada15variancereducedpreference}, and extending dLLMs to long contexts~\citep{liu2025longlladaunlockinglongcontext, he2025ultralladascalingcontextlength} and multimodal settings~\citep{you2025llada}. Despite these advances, the efficiency of dLLMs in long-context scenarios remains underexplored.

\vspace{-2mm}
\subsection{Efficient Inference for dLLMs}
\vspace{-1mm}
In autoregressive LLMs, sparse attention methods such as MInference~\citep{jiang2024minference}, FlexPrefill~\citep{lai2025flexprefill}, XAttention~\citep{xu2025xattention}, and NSA~\citep{yuan2025nativesparseattentionhardwarealigned} reduce long-context attention cost via adaptive or block-sparse patterns, while StreamingLLM~\citep{xiao2024efficientstreaminglanguagemodels}, H2O~\citep{zhang2023h2o}, and SnapKV~\citep{li2024snapkv} compress the KV cache by retaining only important entries. However, these techniques target causal attention where a KV cache is naturally built during left-to-right generation, and do not directly apply to the bidirectional attention in dLLMs where no such cache exists. For dLLMs, Fast-dLLM~\citep{wu2025fastdllmtrainingfreeaccelerationdiffusion} and Fast-dLLM v2~\citep{wu2025fast} introduce KV caching across denoising steps by reusing key-value representations from previous iterations. dKV-Cache~\citep{ma2026dkv} proposes adaptive caching that selectively updates KV entries based on token confidence. SparseD~\citep{wang2025sparsed}, Sparse-dLLM~\citep{song2025sparsedllmacceleratingdiffusionllms}, and d$^2$Cache~\citep{jiang2025d2cacheacceleratingdiffusionbasedllms} exploit inherent attention sparsity for dynamic cache eviction. However, all these methods operate within the standard inference loop where the entire sequence is processed at every step. Our work instead disaggregates the prefix computation from iterative decoding at the system level, and applies sparse chunk retrieval over a static prefix KV cache.
\vspace{-3mm}
\subsection{Prefill-Decode Disaggregation}
\vspace{-1mm}
In autoregressive LLM serving, prefill is compute-bound while decode is memory-bound. DistServe~\citep{zhong2024distserve} exploits this asymmetry by assigning the two phases to separate GPU clusters. Mooncake~\citep{qin2024mooncake} transfers KV caches between prefill and decode nodes via a distributed cache pool, SPAD~\citep{zhang2025spad} designs specialized hardware for each phase, and Semi-PD~\citep{hong2025semi} introduces a hybrid approach with disaggregated computation and unified storage. This principle has not been applied to dLLMs, where every denoising step performs a full forward pass over the entire sequence, making inference compute-bound throughout. Our work bridges this gap by computing the prefix KV cache once and reusing it across all denoising iterations, and further analyzes the potential memory bottleneck introduced by caching.

\section{Preliminary: Masked Diffusion Models}

Masked diffusion language models (dLLMs) define a forward noising process that progressively corrupts a discrete token sequence $\mathbf{x}_0 = (x_1, \ldots, x_L)$ by replacing tokens with a special \texttt{[MASK]} token. At each diffusion timestep $t \in [0, 1]$, each token is independently masked with probability $t$, yielding a noised sequence $\mathbf{x}_t$. The reverse (denoising) process is parameterized by a neural network $p_\theta(\mathbf{x}_0 | \mathbf{x}_t)$ that predicts the original clean tokens given the partially masked input.

During training, the model is optimized to minimize the cross-entropy loss over masked positions:
\begin{equation}
\mathcal{L} = \mathbb{E}_{t, \mathbf{x}_0, \mathbf{x}_t} \left[ -\sum_{i: x_t^i = \texttt{[M]}} \log p_\theta(x_0^i | \mathbf{x}_t) \right]
\end{equation}

During inference, the model starts from a fully masked sequence and iteratively unmasks tokens over $T$ denoising steps. At each step, the model predicts all masked positions simultaneously, and a subset of high-confidence predictions are unmasked according to a scheduling strategy. This parallel decoding enables dLLMs to generate multiple tokens per step, but at each step the model performs full self-attention over the entire sequence (prefix + response), resulting in computational cost that scales with the total length at every iteration.

\vspace{-2mm}
\section{Motivation}
\vspace{-2mm}
\subsection{Lost-in-the-Middle in dLLMs}

\begin{figure}[t]
  \centering
  \includegraphics[width=\columnwidth]{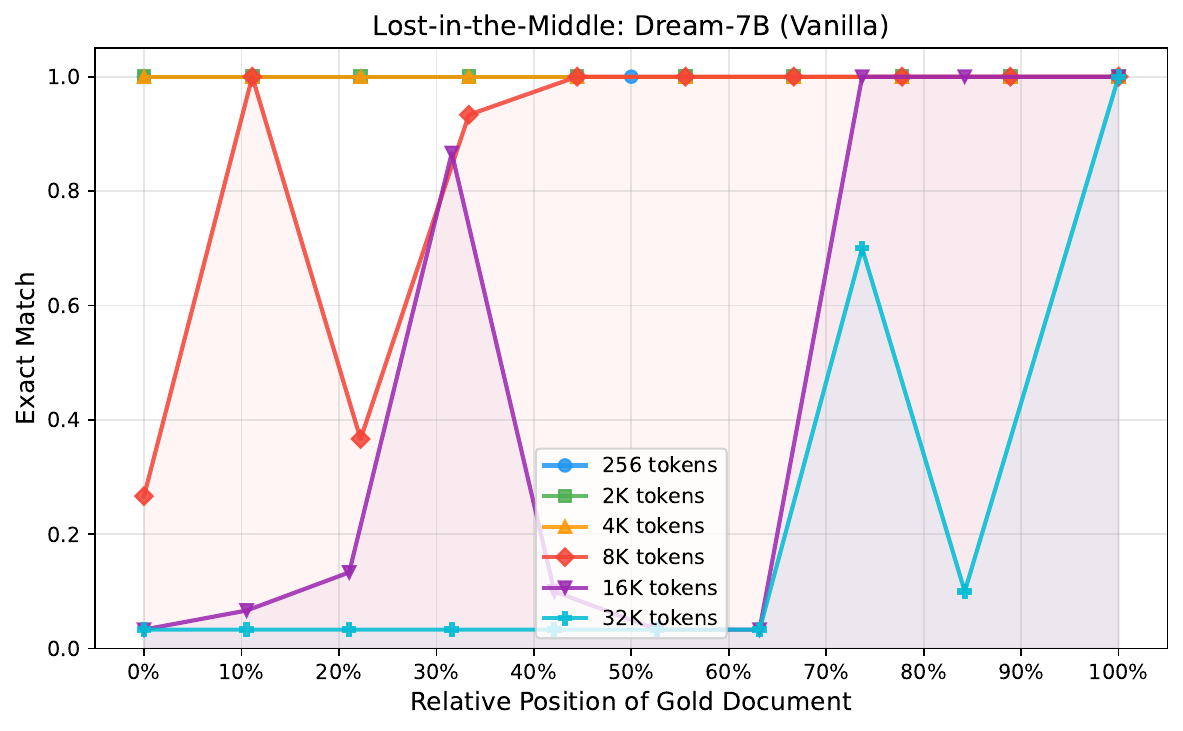}
  \caption{Lost-in-the-Middle evaluation on Dream-7B (training length = 2K). Context extrapolation via YaRN scaling. Native range (256--2K) and YaRN $\times$2 (4K) achieve EM = 1.0 across all positions. YaRN $\times$4 (8K), YaRN $\times$8 (16K), and YaRN $\times$16 (32K) show increasing degradation. Each position is evaluated with 30 samples; 10 evenly spaced positions per context length.}
  \label{fig:lim_extrap}
\end{figure}

Autoregressive LLMs suffer from the ``lost-in-the-middle'' phenomenon~\citep{liu2024lost}, where retrieval accuracy drops for information placed in the middle of the context. We evaluate whether dLLMs share this bias using a position-controlled multi-document QA task and find three key observations: (i) Within the native training range (256--2K tokens) and YaRN $\times$2 extrapolation (4K), Dream-7B achieves perfect accuracy at all positions; (ii) Further extrapolation (8K, 16K, 32K) introduces emerging positional sensitivity (Figure~\ref{fig:lim_extrap}), with accuracy skewing toward positions closer to the response, unlike the U-shaped curve in AR LLMs where both the beginning and end are favored; (iii) In dLLMs, bidirectional attention produces a monotonic decay: tokens near the response receive strong attention regardless of their absolute position, while distant tokens are uniformly neglected. This locality-driven degradation motivates our chunk-based selective retrieval strategy.

\vspace{-2mm}
\subsection{Locality of Attention Decay}

\begin{figure}[t]
  \centering
  \vspace{-2mm}
  \includegraphics[width=\columnwidth]{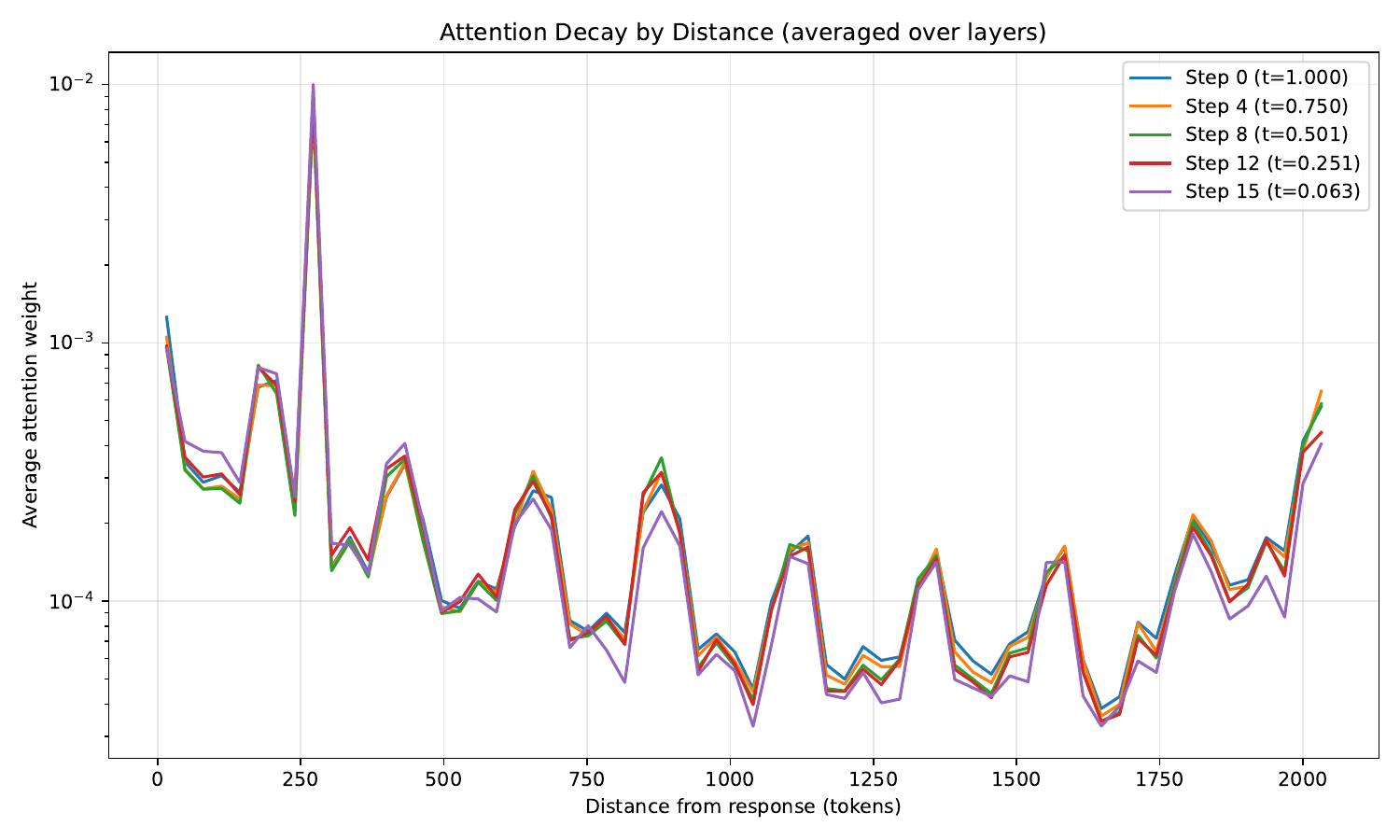}
  \vspace{-2mm}
  \caption{Attention weight decay as a function of distance from response tokens to prefix tokens, measured at different denoising steps. Attention decays rapidly with distance, exhibiting strong locality bias. This decay becomes more pronounced in later denoising steps as token predictions stabilize.}
  \vspace{-2mm}
  \label{fig:attn_decay_step}
\end{figure}

We further analyze the attention patterns of Dream-7B during denoising to understand how response tokens attend to the prefix. We measure the average attention weight from response tokens to prefix tokens as a function of distance (number of tokens separating them). We observe three key findings (Figure~\ref{fig:attn_decay_step}): (i) Attention weights decay rapidly with distance, with response tokens concentrating most of their attention mass on nearby prefix tokens; (ii) The decay becomes more pronounced as denoising progresses and token predictions stabilize, suggesting that full attention over the entire prefix is largely redundant in later steps; (iii) Beyond the overall decay trend, attention exhibits sparse, quasi-periodic spikes at specific prefix positions, with the dominant spike concentrating 25\% of attention mass, stable across all denoising steps and consistent across layers 5--27, corresponding to salient tokens (e.g., segment boundaries).

This locality and sparsity pattern directly motivates our \ours{} design: since distant prefix tokens contribute negligibly to response generation and a small number of chunks capture the majority of useful attention signal, we can cache the prefix KV once with parallel chunk processing and selectively retrieve only relevant chunks during decoding, achieving significant speedups without sacrificing quality.

\begin{figure*}[t]
    \centering
    \includegraphics[width=\textwidth]{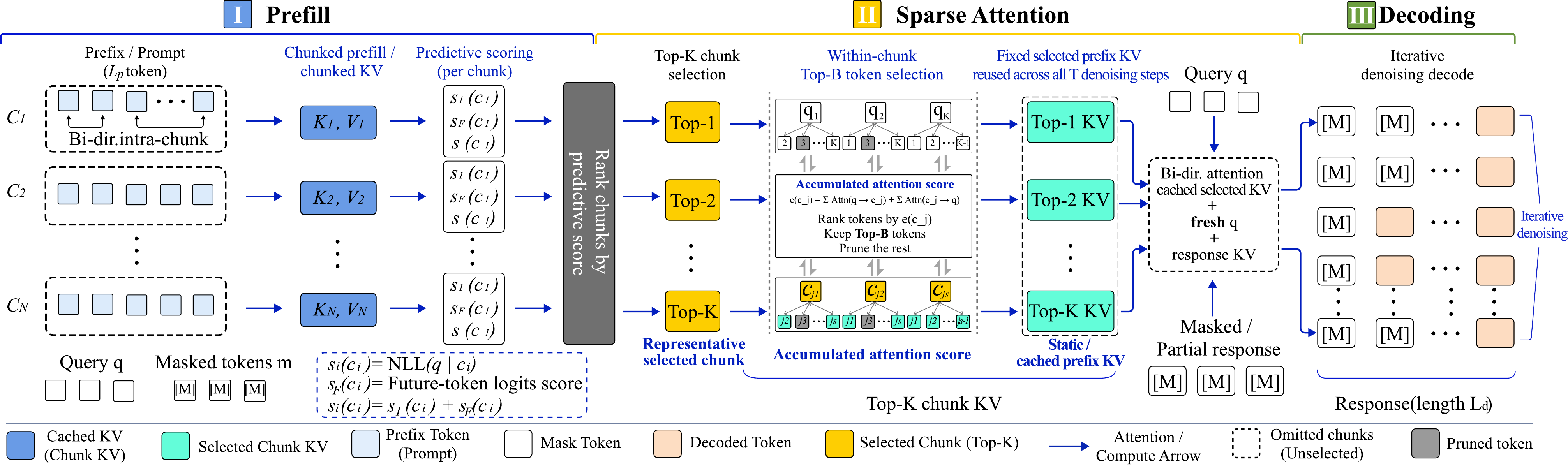}
    \caption{Overview of \ours{}. \textbf{(I) Prefill:} The prefix is partitioned into $N$ chunks, each independently prefilled with intra-chunk attention to produce per-chunk KV caches and predictive and token scores. Chunks are ranked by a predictive score combining a self-information score and a pseudo-label score. After top-$K$ chunk selection and optional top-$B$ token retention, the retained per-chunk KV entries are assembled directly into the reusable prefix KV cache. \textbf{(II) Sparse Attention:} During decoding, only the selected prefix KV entries participate in cross-attention with the response tokens. \textbf{(III) Decoding:} Iterative denoising progressively unmasks the response over $T$ steps, reusing the cached KV without recomputation.}
    \label{fig:pipeline}
\end{figure*}

\section{Method}
We present \ours{}, a two-stage framework that disaggregates prefix computation from iterative denoising. Instead of re-encoding the prefix at every denoising step, we process it once in a prefill stage and cache its KV representations for reuse during decoding. This reduces computational complexity from $O((L_p + L_q + L_d)^2 \cdot T)$ to $O\bigl(N(C+L_q+|\mathbf{m}|)^2 + (L_d^2 + KC)T\bigr)$, where $L_p$ is the prefix length, $L_q$ is the query length, $L_d$ is the number of decoding tokens, $|\mathbf{m}|$ is the pseudo-label length, $T$ is the number of denoising steps, $N = \lceil L_p / C \rceil$ is the number of chunks, and $C$ is the chunk size. The first term accounts for chunk scoring and cache construction with the query and pseudo-label window, while the second term accounts for iterative decoding over the selected prefix cache.
\vspace{-2mm}
\subsection{Prefill}

Inspired by the attention decay observed in Section~\ref{fig:attn_decay_step}, we observe that attending to all $N$ chunks is unnecessary, nor do chunks need to attend to each other. Instead, we propose a predictive prefill strategy that independently processes each chunk, scores its relevance to the query, and selects only the top-$K$ informative chunks for decoding.

\paragraph{Chunk Prefill.} We partition the prefix into $N = \lceil L_p / C \rceil$ non-overlapping chunks $\{\mathbf{c}_1, \ldots, \mathbf{c}_N\}$, each of size $C$ tokens. A special BOS token is prepended to each chunk as a delimiter. We obtain \emph{pseudo-labels} $\mathbf{m}$ by running a single-pass denoising over the query with the full prefix (without chunking) to produce an initial response estimate; these pseudo-labels guide chunk scoring without requiring ground-truth targets. For each chunk, we concatenate it with the query tokens $\mathbf{q}$ and pseudo-labels $\mathbf{m}$ to form $[\mathbf{c}_i; \mathbf{q}; \mathbf{m}]$, and run a bidirectional masked forward pass with positional indices reset from 0 for this chunked input. This yields per-chunk KV caches:
\begin{equation}
\begin{aligned}
\mathbf{K}_i, \mathbf{V}_i
&= \operatorname{KV}_{\theta}\!\left([\mathbf{c}_i;\mathbf{q};\mathbf{m}]\right)[\mathbf{c}_i], \\
\mathbf{K}_i, \mathbf{V}_i
&\in \mathbb{R}^{H \times C \times d}.
\end{aligned}
\end{equation}
Here, $H$ is the number of attention heads and $d$ is the head dimension. Only the KV entries corresponding to the prefix chunk $\mathbf{c}_i$ are retained for later chunk selection. Since chunks are independent, they can be processed in parallel across devices.
\paragraph{Predictive Score.} We score each chunk using two complementary signals obtained during prefill to evaluate its relevance as an \emph{inter-chunk sparsity estimator}. First, we compute the \textit{Self-Information Score} as the negative log-likelihood of the query window $\mathbf{q}$ conditioned on the chunk:
\begin{equation}
s_{\text{I}}(\mathbf{c}_i) =
-\frac{1}{L_q}
\sum_{j=1}^{L_q}
\log p_\theta(q_j \mid \mathbf{c}_i, \mathbf{q}_{\mathrm{mask}(j)})
\end{equation}
where $\mathbf{q}_{\mathrm{mask}(j)}$ denotes the query window with $q_j$ masked. The scoring uses the bidirectional masked attention pattern of the dLLM rather than an autoregressive causal mask. A lower NLL indicates that the chunk provides more information relevant to the query. Second, we compute the \textit{Pseudo-Label Score} using the pseudo-labels $\mathbf{m}$ obtained during prefill. We evaluate how well each chunk predicts these pseudo-labels:
\begin{equation}
s_{\text{P}}(\mathbf{c}_i) = -\frac{1}{|\mathbf{m}|} \sum_{j=1}^{|\mathbf{m}|} \log p_\theta(m_j \mid \mathbf{c}_i, \mathbf{q})
\end{equation}
where $\mathbf{m}$ denotes the pseudo-labels obtained from a preliminary diffusion generation.

\subsection{Sparse Attention}

Our framework introduces sparsity at two levels.
\vspace{-2mm}
\paragraph{Intra-chunk sparsity.} During prefill, each chunk performs bidirectional self-attention only within itself, avoiding the quadratic cost of full-prefix attention. The query tokens participate in bidirectional attention with each chunk and serve as a proxy to select informative entries from each per-chunk KV cache. Specifically, for each token $c_j$ in chunk $\mathbf{c}_i$, we compute its eviction score as the cumulative bidirectional attention weight between the token and the query:
\begin{equation}
\scriptsize e(c_j) = \sum_{k=1}^{L_q} \text{Attn}(q_k, c_j) + \sum_{k=1}^{L_q} \text{Attn}(c_j, q_k)
\end{equation}
Tokens are ranked by eviction score, and only the top-$B$ KV entries per chunk are retained, maintaining a fixed budget while preserving query-relevant information.
\paragraph{Inter-chunk sparsity.} We combine the two length-normalized negative log-likelihood terms with equal weight, $s(\mathbf{c}_i) = s_{\text{I}}(\mathbf{c}_i) + s_{\text{P}}(\mathbf{c}_i)$, and retain the $K$ chunks with the lowest combined scores ($K \ll N$). Length normalization makes the two terms comparable, while equal weighting avoids introducing an additional hyperparameter. The reusable prefix KV cache is assembled directly from the retained KV entries of the selected chunks. During decoding, only these $K$ chunks participate in attention, so the response tokens attend to $K \cdot B$ prefix tokens rather than the full prefix of length $L_p$, significantly reducing the per-step computation.

\subsection{Decoding}

\paragraph{Prefix Reuse.} The retained KV entries of the selected chunks and the query KV representations form a static cache that is reused across all $T$ denoising steps. The query representations are computed once during prefill and remain unchanged during decoding. At each denoising step, only the response tokens are processed, and their KV representations are concatenated with the cached KV of the selected chunks and query tokens. This yields a per-step cost of $O(L_d^2)$ instead of $O((L_p + L_q + L_d)^2)$.
\vspace{-2mm}
\paragraph{Iterative Denoising.} Starting from a fully masked response sequence, the model iteratively unmasks tokens over $T$ denoising steps. At each step, the model predicts all remaining masked positions simultaneously, and tokens whose confidence exceeds a threshold $\tau$ are unmasked. As denoising progresses, the number of masked tokens decreases monotonically until the full response is revealed.

\vspace{-3mm}
\section{Experiments}
\vspace{-2mm}
\subsection{Setup}

\paragraph{Benchmarks.} We evaluate \ours{} on two long-context benchmarks: LongBench~\citep{bai2024longbench}, which covers a set of tasks including single-document QA, multi-document QA, summarization, few-shot learning, synthetic tasks, and code completion with context lengths ranging from 2K to 32K tokens; and InfiniteBench~\citep{zhang2024bench}, which extends to contexts exceeding 100K tokens with tasks such as long-document retrieval, book-level QA, and mathematical reasoning.

\vspace{-2mm}
\subsection{Main Results}
\vspace{-1mm}
\textit{Intra-chunk sparsity can improve performance in dLLMs}, contrary to the performance drop observed in AR LLMs under sparse attention.

\begin{table*}[t]
\centering
\scriptsize
\setlength{\tabcolsep}{2pt}
\renewcommand{\arraystretch}{1.15}
\caption{Performance comparison on LongBench. \textbf{Bold} indicates the best performance among acceleration methods. In sparse variants, we retain the top-$B$ highest-attention tokens per chunk, with $B=512$ in our experiments.}
\label{tab:longbench_results}
\resizebox{\linewidth}{!}{
\begin{tabular}{l*{17}{c}}
\toprule
\multirow{5}{*}{Method} 
& \multicolumn{3}{c}{Single-Doc. QA}
& \multicolumn{3}{c}{Multi-Doc. QA}
& \multicolumn{4}{c}{Summarization}
& \multicolumn{2}{c}{Few-shot Learning}
& \multicolumn{2}{c}{Synthetic}
& \multicolumn{2}{c}{Code}
& \multirow{5}{*}{Ave.\ Score}
\\
\cmidrule(lr){2-4}
\cmidrule(lr){5-7}
\cmidrule(lr){8-11}
\cmidrule(lr){12-13}
\cmidrule(lr){14-15}
\cmidrule(lr){16-17}
& \rotatebox{60}{NarrativeQA}
& \rotatebox{60}{Qasper}
& \rotatebox{60}{MF-en}
& \rotatebox{60}{HotpotQA}
& \rotatebox{60}{2WikiMQA}
& \rotatebox{60}{Musique}
& \rotatebox{60}{GovReport}
& \rotatebox{60}{QMSum}
& \rotatebox{60}{MultiNews}
& \rotatebox{60}{SAMSum}
& \rotatebox{60}{TREC}
& \rotatebox{60}{TriviaQA}
& \rotatebox{60}{PCount}
& \rotatebox{60}{PRe}
& \rotatebox{60}{LCC}
& \rotatebox{60}{RB-P}
& \\
\midrule
\multicolumn{18}{c}{\textit{Dream-v0-Base-7B}~\cite{ye2025dream7bdiffusionlarge} \small{(extrapolated to 128K via YaRN)}}\\
\midrule

YaRN (128K)
& 7.08 & 16.33 & 32.37
& 9.30 & 10.75 & 6.89
& 5.25 & 14.88 & 8.09 & 29.00
& 26.00 & 55.04
& 0.68 & 7.40
& 26.28 & 31.44
& 17.92
\\

Vanilla (2K)
& 20.27 & 19.33 & 40.50
& 29.38 & 20.07 & 15.06
& 5.22 & 17.76 & 8.42 & 33.77
& 60.00 & 81.87
& 0.50 & 16.50
& 46.95 & 41.99
& 28.60
\\
\midrule

Fast-dLLM
& 9.70 & 18.03 & 36.28
& 11.32 & 11.94 & 7.07
& 5.24 & 13.32 & 7.88 & 31.09
& 46.00 & 70.61
& 0.27 & 3.50
& 24.62 & 29.23
& 20.38
\\

Fast-dLLM v2
& \textbf{27.50} & \textbf{33.25} & \textbf{49.76}
& \textbf{46.00} & \textbf{35.92} & \textbf{22.90}
& 4.28 & 17.53 & 6.20 & 31.38
& 43.50 & 78.28
& \textbf{4.00} & \textbf{88.50}
& 10.41 & 7.12
& 31.66
\\

Sparse-dLLM
& 3.19 & 9.45 & 20.25
& 7.04 & 6.11 & 4.99
& 5.27 & 5.92 & \textbf{8.13} & 13.04
& 0.00 & 10.96
& 0.23 & 2.62
& 17.01 & 22.46
& 8.54
\\

dKV-Cache
& 3.57 & 6.60 & 18.33
& 7.02 & 6.34 & 4.44
& 5.32 & 6.07 & 7.84 & 7.71
& 0.00 & 7.17
& 0.47 & 1.60
& 14.74 & 19.92
& 7.32
\\

\rowcolor{mycolor!30}
Ours (inter-sparsity)
& 12.14 & 18.28 & 38.26
& 15.32 & 16.13 & 9.75
& 5.40 & \textbf{17.74} & 7.70 & 29.24
& 46.00 & 70.68
& 0.00 & 13.75
& 21.49 & 30.23
& 22.01
\\

\rowcolor{mycolor!30}
Ours (inter + intra-sparsity)
& 19.90 & 26.72 & 46.40
& 36.22 & 22.10 & 19.96
& \textbf{5.58} & 17.52 & 8.02 & \textbf{32.96}
& \textbf{64.50} & \textbf{92.11}
& 1.54 & 76.02
& \textbf{25.90} & \textbf{57.98}
& \textbf{34.59}
\\

\midrule
\multicolumn{18}{c}{\textit{UltraLLaDA}~\cite{he2025ultralladascalingcontextlength} \small{(native 128K)}}\\
\midrule

Vanilla (128K)
& 17.04 & 22.71 & 33.20
& 15.98 & 16.71 & 12.20
& 5.38 & 17.30 & 8.00 & 37.10
& 79.50 & 90.53
& 1.48 & 96.08
& 67.77 & 54.97
& 36.00
\\
\midrule

Fast-dLLM
& \textbf{22.03} & 21.58 & 37.31
& 17.31 & 16.97 & 11.04
& 5.30 & 17.71 & 8.20 & 33.18
& 79.50 & 90.53
& 1.36 & 94.47
& 66.08 & 53.10
& 35.98
\\

Sparse-dLLM
& 21.30 & 23.16 & 35.43
& 19.08 & 19.27 & \textbf{15.39}
& 5.35 & 17.51 & 8.26 & 35.37
& 77.50 & 91.43
& 0.73 & \textbf{97.58}
& 66.89 & 52.57
& 36.68
\\

dKV-Cache
& 17.86 & 22.36 & 32.99
& 17.89 & 18.39 & 14.25
& \textbf{5.41} & 17.18 & 8.04 & \textbf{36.30}
& \textbf{80.50} & 90.53
& 1.48 & 94.88
& \textbf{67.96} & 54.61
& 36.29
\\

\rowcolor{mycolor!30}
Ours (inter-sparsity)
& 12.27 & 22.40 & \textbf{39.94}
& \textbf{22.21} & \textbf{23.48} & 9.64
& 5.33 & 17.85 & \textbf{8.48} & 34.33
& 67.00 & 88.16
& 2.38 & 90.91
& 65.10 & \textbf{62.67}
& 35.64
\\

\rowcolor{mycolor!30}
Ours (inter + intra-sparsity)
& 17.07 & \textbf{24.71} & 36.07
& 21.57 & 17.87 & 14.30
& 5.21 & \textbf{18.12} & 8.46 & 34.43
& 77.50 & \textbf{92.50}
& \textbf{2.50} & 96.91
& 66.92 & 58.20
& \textbf{37.02}
\\

\bottomrule
\end{tabular}}
\end{table*}

\vspace{-2mm}
\paragraph{LongBench.}
Table~\ref{tab:longbench_results} presents the performance comparison on LongBench. We highlight several observations: (i) \ours{} achieves state-of-the-art performance on both Dream-7B and UltraLLaDA, outperforming the strongest baseline Fast-dLLM v2 by 2.93 points on Dream-7B and mitigating the lost-in-the-middle phenomenon on UltraLLaDA; (ii) \ours{} obtains the best results on 9 out of 16 subtasks, with particularly large gains on retrieval-intensive tasks such as Synthetic PRe (78.93 vs.\ 16.71), demonstrating that inter-chunk sparsity effectively identifies relevant context; (iii) Adding intra-chunk sparsity (Ours, inter + intra) achieves state-of-the-art performance while further reducing computation.
\vspace{-2mm}
\paragraph{InfiniteBench.} We evaluate \ours{} on InfiniteBench with contexts exceeding 128K tokens. Results are presented in Table~\ref{tab:infinitebench_results}. On Dream-7B, Ours (inter + intra-sparsity) achieves 43.62 average accuracy, surpassing the strongest baseline Fast-dLLM v2 (30.32) by over 13 points, with particularly strong gains on Passkey (95.42) and Number retrieval (70.00), showing that predictive context selection can improve long-context performance without additional training.

\begin{table}[t]
\centering
\scriptsize
\setlength{\tabcolsep}{2pt}
\renewcommand{\arraystretch}{1.15}
\caption{Performance comparison on InfiniteBench. Accuracy is reported as percentage. \textbf{Bold} indicates the best performance among acceleration methods. In sparse variants, we retain the top-$B$ highest-attention tokens per chunk, with $B=512$ in our experiments.}
\label{tab:infinitebench_results}
\resizebox{\columnwidth}{!}{%
\begin{tabular}{l*{7}{c}}
\toprule
Method
& Passkey
& Number
& KV
& LongBook
& Math
& Code
& Ave. \\
\midrule
\multicolumn{8}{c}{\textit{Dream-v0-Base-7B}~\cite{ye2025dream7bdiffusionlarge} \small{(extrapolated to 128K via YaRN)}}\\
\midrule
YaRN (128K)
& 0.00 & 0.00 & 0.00 & 6.11 & 3.43 & 7.87 & 2.90 \\
Vanilla (2K)
& 1.69 & 2.54 & 0.00 & 35.81 & 34.57 & 0.00 & 12.44 \\
\midrule
Fast-dLLM
& 0.17 & 3.05 & 0.00 & 8.73 & 3.43 & 8.38 & 3.96 \\
Fast-dLLM v2
& 27.12 & 27.12 & 0.00 & 57.21 & 24.29 & 46.19 & 30.32 \\
Sparse-dLLM
& 39.32 & 25.59 & 0.00 & 1.31 & 3.43 & 3.81 & 12.24 \\
dKV-Cache
& 0.51 & 0.00 & 0.00 & 5.68 & 4.00 & 3.05 & 2.21 \\
\rowcolor{mycolor!30}
Ours (inter + intra-sparsity)
& \textbf{95.42} & \textbf{70.00} & 0.00 & 53.71 & 24.57 & \textbf{18.02} & \textbf{43.62} \\
\midrule
\multicolumn{8}{c}{\textit{UltraLLaDA}~\cite{he2025ultralladascalingcontextlength} \small{(native 128K)}}\\
\midrule
Vanilla (128K)
& 12.99 & 2.45 & 0.00 & 8.86 & 10.43 & 0.00 & 5.79 \\
\midrule
Fast-dLLM
& \textbf{16.10} & 6.61 & 0.00 & 7.86 & 14.29 & 0.25 & 7.52 \\
Sparse-dLLM
& 8.14 & 2.37 & 0.00 & 5.24 & 7.43 & \textbf{1.52} & 4.12 \\
dKV-Cache
& 5.82 & 6.75 & 0.00 & 8.30 & 16.00 & 0.00 & 6.15 \\
\rowcolor{mycolor!30}
Ours (inter + intra-sparsity)
& 13.22 & \textbf{30.00} & 0.00 & \textbf{14.41} & \textbf{26.00} & \textbf{1.52} & \textbf{14.19} \\
\bottomrule
\end{tabular}%
}
\end{table}

\noindent All evaluated dLLM methods obtain zero scores on the InfiniteBench KV retrieval task. To verify that this result is not an artifact of our evaluation pipeline, we additionally evaluate long-context AR baselines based on LLaMA3-8B-Instruct: NTK-Aware obtains 9.40 and ChunkLlama obtains 3.60. These low scores indicate that InfiniteBench KV retrieval remains difficult for current long-context methods more broadly, rather than pointing to a failure specific to dLLM evaluation or our context-selection pipeline.

\vspace{-2mm}
\subsection{Efficiency Analysis}
\vspace{-1mm}
\textit{We show that \ours{} scales sub-linearly with length via fixed-size chunk selection, surpassing the strongest baseline Sparse-dLLM at 16K and 32K.}

\begin{figure}[t]
    \centering
    \includegraphics[width=\columnwidth]{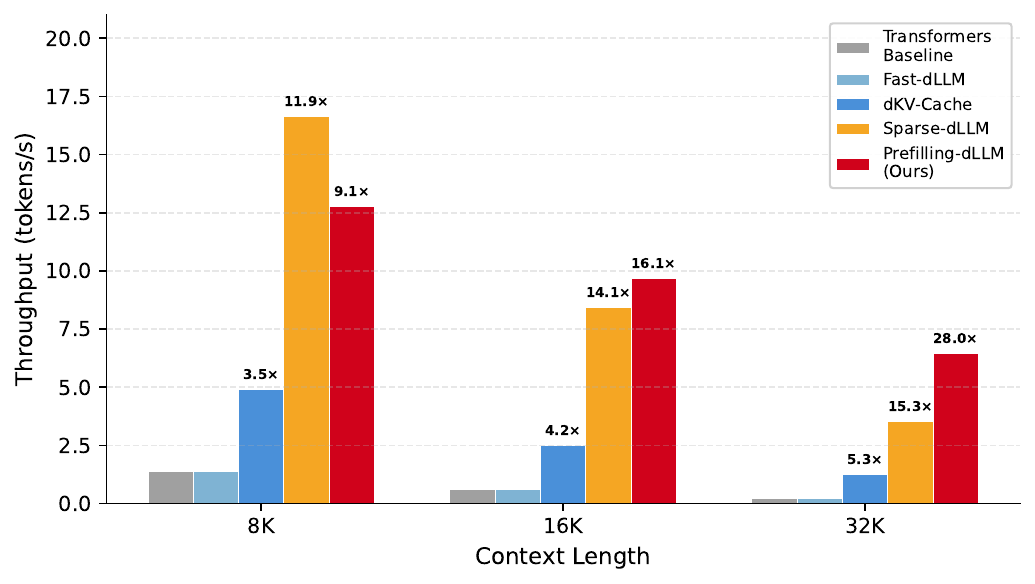}
    \caption{Throughput comparison (tokens/s) on LongBench MF-en at varying context lengths (Dream-7B, bf16, GQA with 32 query heads, 8 KV heads, head dim 128, single A100 GPU, 32 generated tokens, 5 measured samples). Labels above bars show speedup relative to the Transformers baseline.}
    \label{fig:efficiency}
\end{figure}

\noindent All wall-clock results are measured end-to-end. They include pseudo-label generation, chunk scoring, KV/cache construction, data movement, and diffusion decoding.

\begin{figure}[t]
    \centering
    \includegraphics[width=\columnwidth]{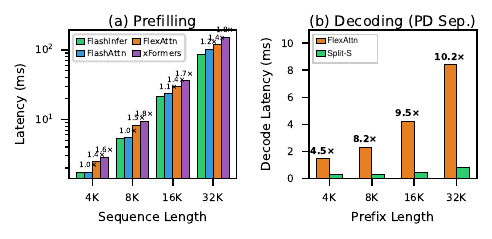}
    \caption{Attention kernel benchmark (bf16, GQA 32/8 heads, single A100). \textbf{(a) Prefilling:} FlashInfer achieves the lowest latency; Flash Attention is 1.0--1.2$\times$ slower; FlexAttention adds 1.4--1.5$\times$ overhead; xFormers is 1.6--1.8$\times$ slower. Labels show relative slowdown vs.\ FlashInfer. \textbf{(b) Decoding (PD separation, $L_d{=}32$, $S{=}4$ splits):} After top-$K$ selection, Split-S partitions the resulting cached KV sequence into $S$ compute shards and processes them in parallel via the batch dimension.}
    \label{fig:attn_kernel}
\end{figure}

\noindent As shown in Figure~\ref{fig:efficiency}, we highlight several key observations: (i) \ours{} achieves increasing speedups as context grows (9.1$\times$ at 8K, 16.1$\times$ at 16K, 28.0$\times$ at 32K), because it compresses the context to a fixed budget (top-$K=4$ chunks $\times$ 1024 tokens $\approx$ 4K) regardless of input length, while all baselines must process the full context at every denoising step; (ii) Sparse-dLLM is fastest at 8K (16.62 tok/s) through aggressive token eviction, but degrades rapidly at longer contexts (3.51 tok/s at 32K) because its eviction ratio is fixed; (iii) In contrast, \ours{} surpasses Sparse-dLLM at both 16K and 32K, demonstrating that retrieval-augmented generation provides better quality (Table~\ref{tab:longbench_results}) and superior scaling efficiency.
\vspace{-3mm}
\paragraph{Attention Kernel Comparison.} \textit{\ours{} adopts FlexAttention with Split-S parallelism for decoding, achieving up to 10.2$\times$ speedup over vanilla FlexAttention.} We benchmark attention kernel options for the two phases of PD-separated dLLM inference. For prefilling, we compare Flash Attention~\citep{dao2022flashattention}, FlexAttention~\citep{he2024flexattention}, xFormers Attention~\citep{rabe2021self}, and FlashInfer~\citep{ye2025flashinfer}. As shown in Figure~\ref{fig:attn_kernel}(a), FlashInfer and Flash Attention achieve the lowest prefilling latency, while FlexAttention adds 1.4--1.5$\times$ overhead from block mask evaluation and xFormers is 1.6--1.8$\times$ slower.

For decoding under PD separation, each denoising step computes attention for $L_d$ decoding tokens against a KV sequence of length $L_p + L_q + L_d$, creating a highly asymmetric pattern ($L_d \ll L_p$). FlexAttention parallelizes only along the query dimension, severely underutilizing the GPU. After the top-$K$ chunks are selected, we partition the resulting cached KV sequence into $S$ compute shards solely for parallel attention execution; we refer to this scheme as \textbf{Split-S}. These shards are independent of the prefix chunks used for scoring and retrieval. Attention over the shards is computed in parallel, and the partial outputs are merged using log-sum-exp reduction, yielding a 5.8--10.2$\times$ speedup (Figure~\ref{fig:attn_kernel}b).

\begin{tcolorbox}[colback=findingbg, colframe=gray!60, boxrule=0.5pt, arc=2pt, left=4pt, right=4pt, top=3pt, bottom=3pt]
\textbf{Finding 1:} \emph{Under PD separation, splitting the already selected prefix KV into independent compute shards improves parallelism for short decoding sequences. Processing these shards in parallel and merging their outputs yields a 5.8--10.2$\times$ latency reduction.}
\end{tcolorbox}
\vspace{-2mm}

\vspace{-2mm}
\subsection{Lost-in-the-Middle}
\vspace{-1mm}

As shown in Section~\ref{fig:lim_extrap}, dLLMs exhibit positional sensitivity under context extrapolation, with retrieval accuracy degrading for information placed in the middle of long contexts. We investigate whether \ours{} mitigates this effect by evaluating on the same position-controlled multi-document QA task~\citep{liu2024lost}. Since \ours{} selects the most relevant chunks via predictive scoring rather than relying on positional proximity, we hypothesize that it can attend to informative tokens regardless of their position in the prefix.

\begin{figure}[t]
    \centering
    \includegraphics[width=\columnwidth]{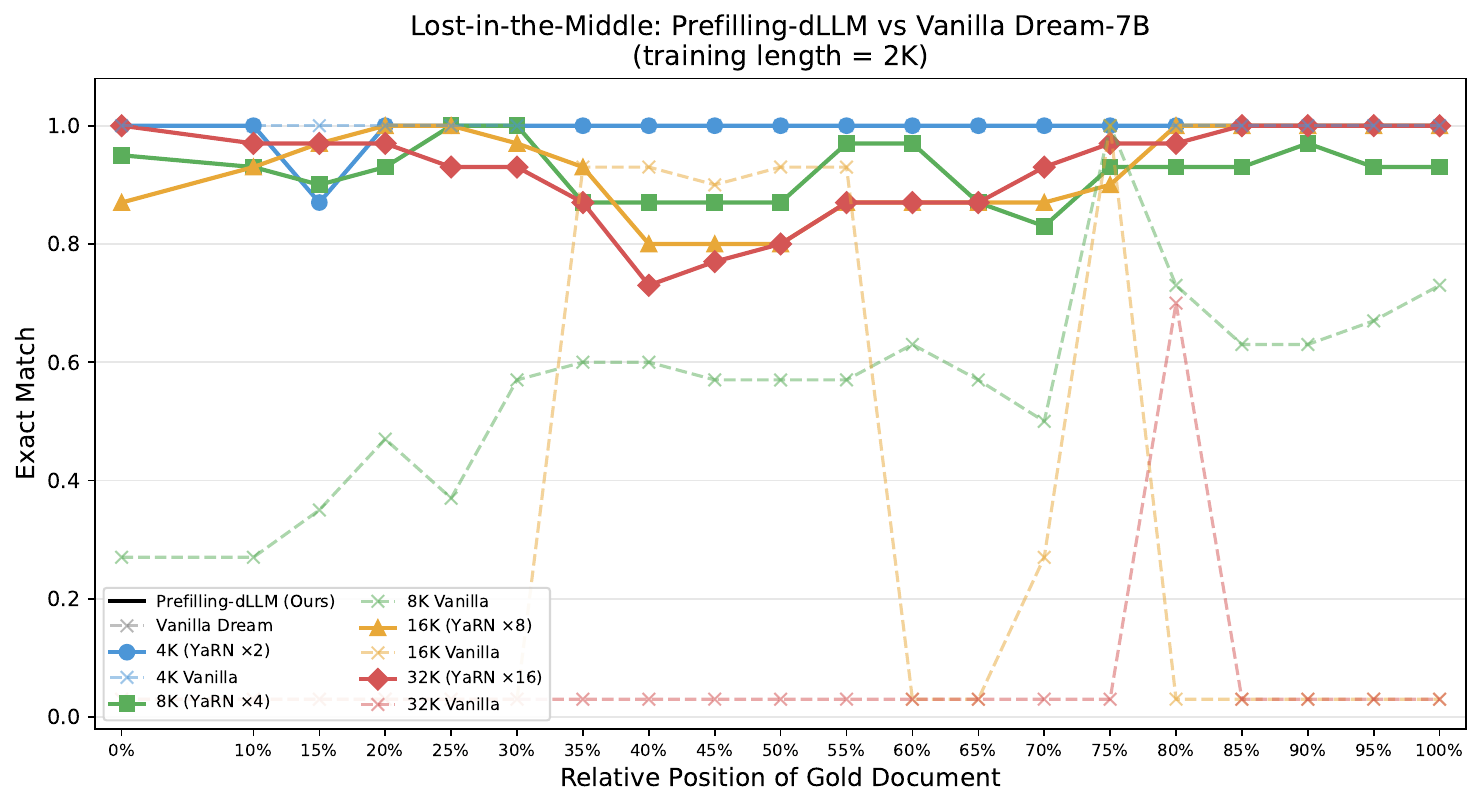}
    \caption{Lost-in-the-Middle evaluation on Dream-7B (training length = 2K). We compare \ours{} (solid) against Vanilla Dream (dashed) with YaRN extrapolation across 4K--32K contexts, measuring exact-match accuracy as a function of gold document position. \ours{} maintains consistently high EM across all positions and context lengths, while Vanilla collapses at 16K and 32K.}
    \label{fig:lim_prefilling_dllm}
\end{figure}
\vspace{-1pt}
\begin{tcolorbox}[colback=findingbg, colframe=gray!60, boxrule=0.5pt, arc=2pt, left=4pt, right=4pt, top=3pt, bottom=3pt]
\textbf{Finding 2:} \emph{\ours{} leverages periodic attention spikes from chunk-level BOS tokens to eliminate the lost-in-the-middle phenomenon in dLLMs.}
\end{tcolorbox}
\vspace{-1mm}
\noindent The periodic attention spikes that cause positional bias in Vanilla inference become the signal that \ours{} leverages for position-invariant chunk retrieval, transforming catastrophic failure into mild degradation at 32K.

\vspace{-3mm}
\subsection{Attention Sink Analysis}
\vspace{-2mm}

\textit{Do the periodic attention spikes from chunk-level BOS tokens degenerate into attention sinks~\citep{xiao2024efficientstreaminglanguagemodels}?} We investigate whether they absorb disproportionate attention mass and bias chunk selection toward positional artifacts.

We analyze the attention patterns during generation for both \ours{} and Vanilla Dream (YaRN $\times$4, 8K context), measuring the fraction of attention mass absorbed by the first-1 token, first-5 tokens, and all BOS tokens across all 28 layers. Both conditions use the full 8K context without chunk selection: Vanilla Dream processes the flat token sequence, while \ours{} segments it into 8 chunks of 1024 tokens with a BOS delimiter prepended to each chunk.

\begin{figure}[t]
    \centering
    \includegraphics[width=\columnwidth]{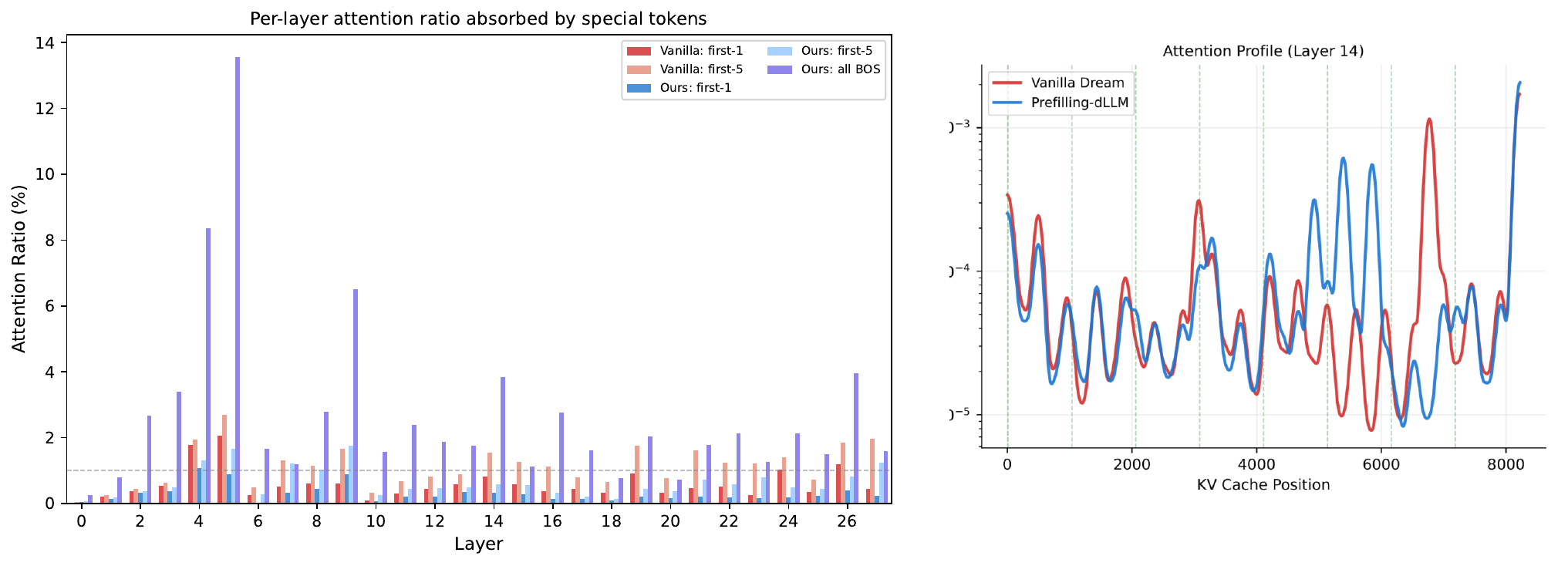}
    \caption{Attention sink analysis (8K context, YaRN $\times$4). \textbf{Left:} Per-layer attention ratio absorbed by the first-1, first-5, and all chunk BOS tokens; both methods stay below 1\% on average, with the BOS token absorbing only 0.59\% (Vanilla) and 0.30\% (\ours{}). \textbf{Right:} Attention profile at layer 14 (log scale); green dashes mark chunk BOS positions, showing periodic attention spikes that distribute mass uniformly.}
    \label{fig:attention_sink}
\end{figure}

\begin{figure}[t]
    \centering
    \includegraphics[width=0.48\columnwidth]{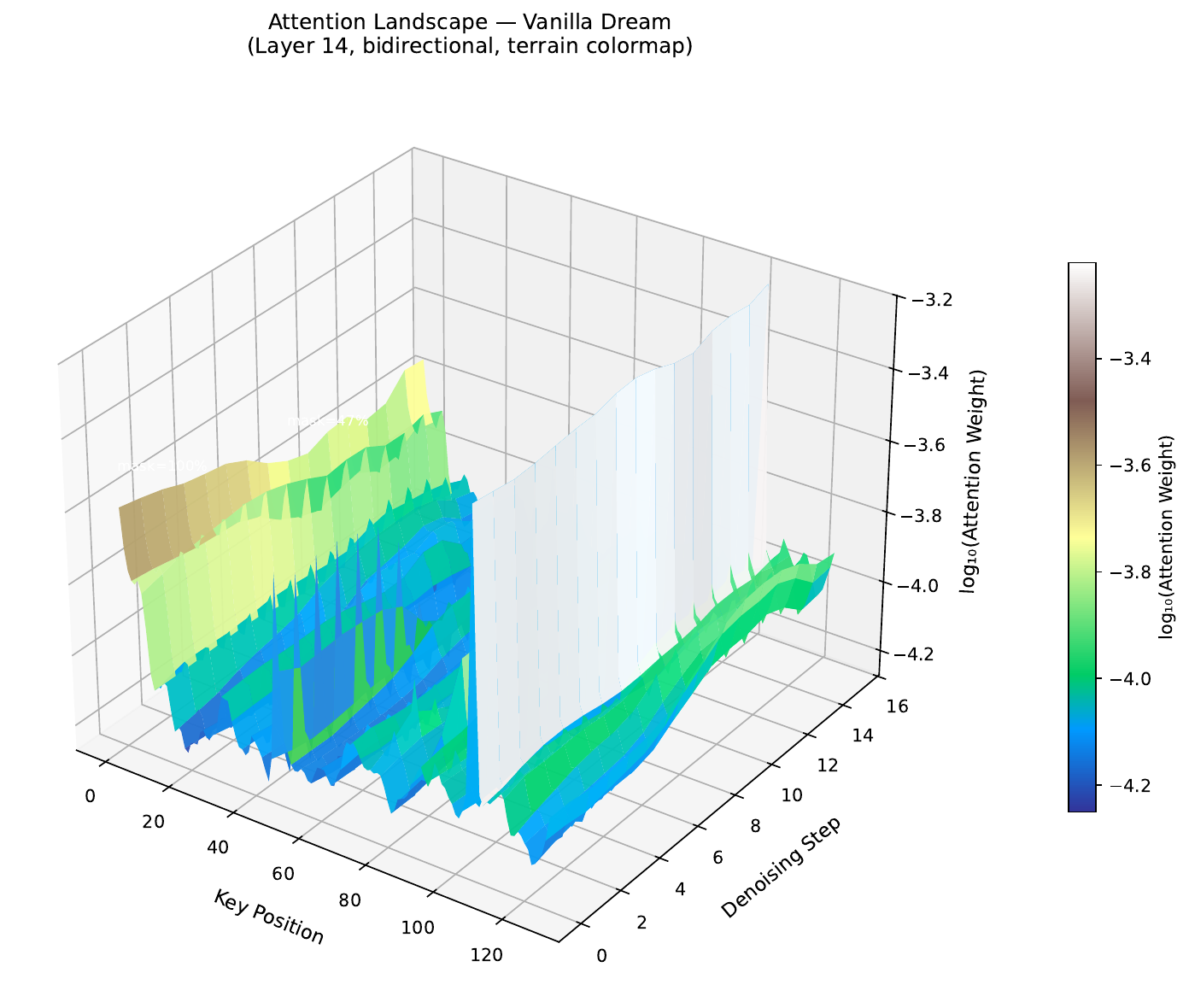}
    \includegraphics[width=0.48\columnwidth]{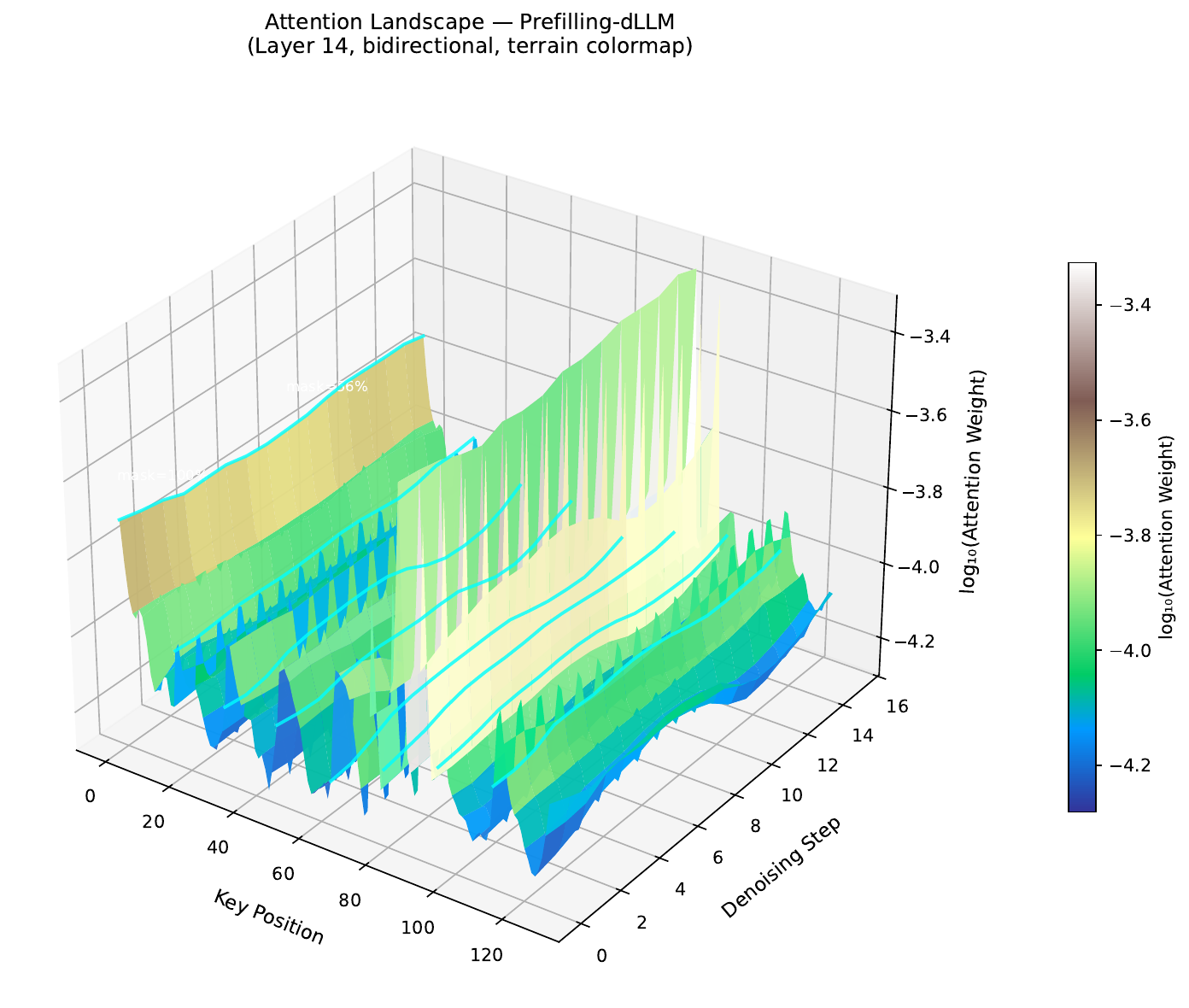}
    \caption{Attention landscape during denoising (layer 14, log-scale). \textbf{Left:} Vanilla Dream shows a flat landscape with mild elevation at the sequence start. \textbf{Right:} \ours{} exhibits periodic ridges (cyan lines) at chunk BOS positions, serving as stable attention anchors without forming dominant sinks.}
    \vspace{-4mm}
    \label{fig:attention_terrain}
\end{figure}

\begin{tcolorbox}[colback=findingbg, colframe=gray!60, boxrule=0.5pt, arc=2pt, left=4pt, right=4pt, top=3pt, bottom=3pt]
\textbf{Finding 3:} \emph{BOS tokens prepended to each chunk in \ours{} act as periodic attention anchors that distribute attention mass uniformly across the context, mitigating the lost-in-the-middle phenomenon.}
\end{tcolorbox}
\vspace{-1mm}
\noindent We highlight several observations from Figure~\ref{fig:attention_sink} and Figure~\ref{fig:attention_terrain}: (i) Unlike AR LLMs where the BOS token absorbs 20--60\% of attention mass~\citep{xiao2024efficientstreaminglanguagemodels}, the first token in dLLMs absorbs only 0.59\% (Vanilla) and 0.30\% (\ours{}) on average across layers; (ii) Even when summing over all 9 chunk BOS tokens, the total BOS attention in \ours{} is only 2.72\%, confirming that chunk BOS tokens serve as segment delimiters without becoming parasitic attention sinks; (iii) As shown in Figure~\ref{fig:attention_terrain}, Vanilla Dream develops a mild ridge only at the sequence start, while \ours{} exhibits periodic ridges at chunk BOS positions that remain stable throughout denoising without growing into dominant peaks.

\begin{tcolorbox}[colback=findingbg, colframe=gray!60, boxrule=0.5pt, arc=2pt, left=4pt, right=4pt, top=3pt, bottom=3pt]
\textbf{Finding 4:} \emph{Chunk BOS tokens form stable, low-magnitude attention ridges throughout denoising, acting as distributed anchors rather than accumulating into dominant sinks.}
\end{tcolorbox}
\vspace{-2mm}
\subsection{Effect of Chunk Size on Prefilling}
\vspace{-1mm}
\begin{figure}[t]
    \centering
    \includegraphics[width=0.85\columnwidth]{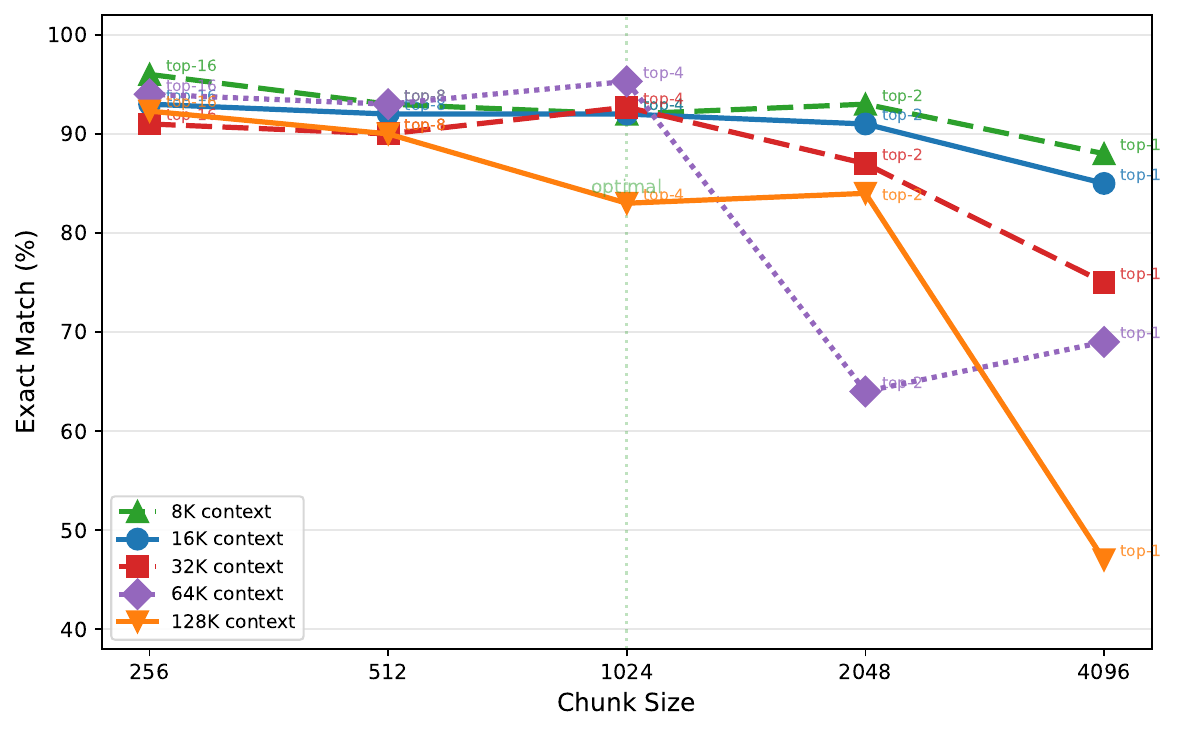}
    \caption{Effect of chunk size on Lost-in-the-Middle retrieval accuracy. We fix the total token budget (top-$K$ $\times$ chunk\_size $=$ 4096) and vary the chunk granularity across 8K--128K contexts. The base model is Vanilla Dream-7B with a 2K training length; all longer contexts require extrapolation. Smaller chunks (256--512) maintain $>$90\% EM even at 128K ($\times$64 extrapolation), while large chunks (4096, top-$K=1$) degrade sharply beyond 32K.}
    \label{fig:chunk_size_ablation}
\end{figure}

We investigate how chunk size affects downstream task performance under a fixed token budget. Specifically, we keep the total number of selected tokens constant at 4096 (i.e., top-$K$ $\times$ chunk\_size $=$ 4096) and vary the chunk granularity across 256, 512, 1024, 2048, and 4096 tokens. Figure~\ref{fig:chunk_size_ablation} shows the results on the Lost-in-the-Middle benchmark at 8K, 16K, 32K, 64K, and 128K context lengths.

The results reveal a clear trade-off that intensifies with context length. At 8K, smaller chunks (256 tokens, top-$K=16$) achieve the highest accuracy (96.0\% EM). As context grows to 16K--64K, chunk size 1024 (top-$K=4$) becomes optimal (92.0\%, 92.7\%, and 95.3\% EM for 16K, 32K, and 64K respectively). At 128K ($\times$64 extrapolation), finer granularity becomes essential: chunk size 256 (top-$K=16$) achieves 92.3\% EM, while chunk size 1024 drops to 83.0\%. Too-large chunks (4096 tokens, top-$K=1$) degrade sharply beyond 32K (47.0\% at 64K, 68.0\% at 128K), confirming that multi-chunk prefilling is essential for long-context dLLM inference.

\begin{tcolorbox}[colback=findingbg, colframe=gray!60, boxrule=0.5pt, arc=2pt, left=4pt, right=4pt, top=3pt, bottom=3pt]
\textbf{Finding 5:} \emph{Chunk size creates a quality--efficiency trade-off: smaller chunks improve retrieval accuracy but underutilize GPU compute. Multi-chunk selection is essential, as single-chunk prefilling fails at all long contexts.}
\end{tcolorbox}
\vspace{-1mm}
\noindent See Appendix~\ref{sec:appendix_mfen_ablation} for additional ablations.

\vspace{-1mm}
\section{Conclusion}
\vspace{-1mm}
We presented \ours{}, a prefill-decode disaggregation framework for dLLMs that caches chunked prefix KV once and selectively retrieves the top-$K$ chunks for decoding, reducing complexity from $O((L_p+L_q+L_d)^2 \cdot T)$ to $O\bigl(N(C+L_q+|\mathbf{m}|)^2 + (L_d^2 + KC)T\bigr)$ with 9.1--28.0$\times$ end-to-end speedup at 8K--32K contexts. Our analysis reveals that chunk-level BOS tokens act as periodic attention anchors that eliminate the lost-in-the-middle phenomenon, and that finer-grained multi-chunk prefilling enables extrapolation to 128K tokens with over 92\% accuracy.

\clearpage
\section*{Limitations}

Our chunk selection is static: the top-$K$ chunks are fixed after prefill with no dynamic re-selection during decoding, so inaccurate pseudo-labels may cause relevant context to be missed. The chunk size $C$ and $K$ must be selected to match the available selected-token budget, as smaller chunks improve accuracy but underutilize GPU compute. Our SCBench results further indicate that static selection is stronger for exact-evidence retention than aggregation-heavy summarization: BM25 + Fast-dLLM v1 attains higher summary F1 under the same selected-context budget. InfiniteBench KV retrieval also remains difficult for both dLLMs and long-context AR baselines, suggesting a broader limitation of current long-context methods. Additionally, FlexAttention lacks paged memory management, requiring the prefix KV cache to be reloaded at every decoding step. Finally, we evaluate only on Dream-7B and UltraLLaDA with English benchmarks; generalization to other dLLM architectures, larger scales, or multilingual settings remains to be verified.

\section*{Use of AI Assistants}
We used AI writing assistants solely for language polishing and proofreading. All research ideas, experimental design, implementation, and scientific conclusions are entirely the authors' own work.

\section*{Acknowledgments}
This project is supported in part by the Theme-based Research Scheme (TRS) projects T45-701/22-R and T41-517/25-N, and GRF Project 17203224 of the Research Grants Council (RGC), Hong Kong SAR, and in part by the AVNET-HKU Emerging Microelectronics \& Ubiquitous Systems (EMUS) Lab.


\clearpage
\appendix

\section*{Appendix}

\section{Implementation Details}
\label{sec:appendix_implementation}

We implement \ours{} on top of three dLLM \emph{base models}: Dream-7B~\citep{ye2025dream7bdiffusionlarge}, UltraLLaDA~\citep{he2025ultralladascalingcontextlength}, and Fast-dLLM v2~\citep{wu2025fast}. We use $T$ denoising steps and choose the chunk size $C$ and the number of chunks $K$ to match fixed selected-token budgets, as detailed below. Unless otherwise specified, all experiments are conducted on NVIDIA A100 GPUs.

\paragraph{Settings.}
All \ours{} runs use bfloat16 inference and greedy decoding with temperature 0. During prefill, we prepend a BOS token to every prefix chunk and score each chunk independently with the query and optional pseudo-label window under a bidirectional masked attention pattern. Unless otherwise stated, selected chunks are cached with \texttt{full-mask} KV construction, continuous chunk positions, and query positions placed after the selected chunks. For sparse variants with intra-chunk sparsity, we retain the top-$B$ highest-scoring tokens per selected chunk and set $B=512$ in the main experiments.
\paragraph{Hyperparameter Selection.}
We choose the chunk size $C$, the number of selected chunks $K$, and the retained-token budget $B$ according to fixed selected-token budgets for each model and context setting. These hyperparameters are not tuned separately for individual test tasks, and no test-set examples are used for hyperparameter selection. Baseline methods are evaluated with their official or recommended configurations whenever available, using the same prompt formatting and truncation protocol described above.
\paragraph{Prompt Formatting and Context Truncation.}
For LongBench, we use the original task-specific prompt templates provided by the LongBench evaluation configuration, where each prompt is rendered by filling the \texttt{\{context\}} and \texttt{\{input\}} fields. For InfiniteBench, we use the raw benchmark-style prompts for each task, following the same structure of an instruction prefix, the long context, and a task query. In both benchmarks, \ours{} separates the rendered prompt into three parts, namely the instruction prefix, the long-context field, and the query suffix. Only the long-context field is partitioned into chunks for chunk scoring, top-$K$ retrieval, and optional top-$B$ token retention; the instruction prefix and query suffix are kept outside the chunk pool and are always included in decoding.

For vanilla and acceleration baselines that cannot process the full prompt within their context window, we apply context-only head-tail truncation: after rendering the same prompt template, we allocate a prompt budget of \texttt{max\_length} minus the generation length, keep the instruction prefix and query suffix unchanged, and truncate only the long-context field by preserving equal-length head and tail portions while dropping the middle. This avoids removing task instructions or the question. For Dream-7B experiments that use YaRN extrapolation, the effective context window is expanded by the corresponding RoPE scaling factor; for UltraLLaDA, we use its native 128K context window. UltraLLaDA main-table experiments are evaluated without a chat template.

\begin{tcolorbox}[colback=findingbg, colframe=gray!60, boxrule=0.5pt, arc=2pt, left=4pt, right=4pt, top=3pt, bottom=3pt]
\small
\textbf{Prompt example.}
For LongBench MF-en, the rendered prompt has the following structure:
\begin{quote}
\footnotesize\ttfamily\raggedright
Read the following text and answer briefly.\\[1mm]
Title: City Archive Renovation\\
The archive reopened after a two-year renovation. The director noted that the new reading room would hold letters, maps, and oral histories from local residents.\\
...\\
In a later interview, the curator said the first public exhibit would focus on neighborhood transit records from 1912.\\[1mm]
Now, answer the following question based on the above text, only give me the answer and do not output any other words.\\[1mm]
Question: [question]\\
Answer:
\end{quote}
\end{tcolorbox}

\paragraph{Dream-v0-Base-7B.}
Dream-v0-Base-7B has a native 2K context window, so the 128K long-context rows use YaRN extrapolation with rope scale factor 64. For \ours{} with inter-chunk sparsity, we use chunk size $C=1024$ and select top-$K=4$ chunks. Chunk ranking uses pseudo-label scoring with 4 pseudo-label tokens and one denoising step. The main intra-chunk sparsity setting keeps the same chunk-selection configuration and adds top-$B$ token retention with $B=512$, using the bidirectional query--chunk attention score for token importance.

\paragraph{UltraLLaDA.}
UltraLLaDA supports native 128K context length. We evaluate it without a chat template in the main tables. For the inter-chunk sparsity setting, we use $C=1024$, select top-$K=2$ chunks, and rank chunks by self-information scoring with a query window of 64 tokens. For the intra-chunk sparsity setting, we use $C=1024$, select top-$K=8$ chunks, and retain top-$B=512$ tokens per chunk with the bidirectional query--chunk attention score.

\paragraph{Computational Budgets and Evaluation Settings.}
Table~\ref{tab:cost_fairness} summarizes the computational budgets and fixed configurations used for the long-context experiments in Tables~\ref{tab:longbench_results} and~\ref{tab:infinitebench_results}, including the prefix length processed during chunk scoring and the selected-prefix token budget retained for decoding. For the timing configuration in Figure~\ref{fig:efficiency}, Dream uses 32 generated tokens, 16 pseudo-label tokens for chunk scoring, and five measured samples. These wall-clock measurements are end-to-end and include scoring, KV cache construction, data movement, and decoding.

\begin{table*}[!t]
\centering
\scriptsize
\setlength{\tabcolsep}{2.5pt}
\renewcommand{\arraystretch}{1.1}
\caption{Computational budgets and evaluation settings for the long-context results in Tables~\ref{tab:longbench_results} and~\ref{tab:infinitebench_results}. ``Context scored'' is the actual prefix processed for selection; ``decode budget'' is the selected prefix used during generation.}
\label{tab:cost_fairness}
\resizebox{\textwidth}{!}{%
\begin{tabular}{lccccccccc}
\toprule
Setting & Context scored & Decode budget & $C$ & top-$K$ & top-$B$ & Gen. & Pseudo labels & YaRN \\
\midrule
Full-context baselines & Reported context & Full context & -- & -- & -- & 32 & -- & Dream: yes; Ultra: no \\
\midrule
Dream inter & 128K, once & $4{\times}1024=4096$ & 1024 & 4 & -- & 32 & 4 / 1 step & Yes ($\times64$) \\
Dream inter + intra & 128K, once & $4{\times}512=2048$ & 1024 & 4 & 512 & 32 & 4 / 1 step & Yes ($\times64$) \\
UltraLLaDA inter & 128K, once & $2{\times}1024=2048$ & 1024 & 2 & -- & 32 & -- & No \\
UltraLLaDA inter + intra & 128K, once & $8{\times}512=4096$ & 1024 & 8 & 512 & 32 & -- & No \\
\bottomrule
\end{tabular}%
}
\end{table*}

\section{Baselines}
\label{sec:appendix_baselines}

We compare \ours{} against standard dLLM inference (full-attention at every denoising step) and dLLM acceleration methods, including Fast-dLLM~\citep{wu2025fastdllmtrainingfreeaccelerationdiffusion}, Fast-dLLM v2~\citep{wu2025fast}, dKV-Cache~\citep{ma2026dkv}, and Sparse-dLLM~\citep{song2025sparsedllmacceleratingdiffusionllms}. We also include YaRN~\citep{peng2023yarn} as a context extrapolation baseline, which extends the native context window of Dream-7B (2K) to 128K via RoPE scaling.

\section{Detailed Ablation Analysis}
\label{sec:appendix_mfen_ablation}

\textit{The ablations show that \ours{} benefits most from selecting a small set of informative chunks, using short pseudo-labels for chunk scoring, and retaining a moderate top-$B$ token budget under intra-chunk sparsity.}

We provide detailed MF-en ablations in Tables~\ref{tab:dream_mfen_ablation}--\ref{tab:ultrallada_mfen_eviction_ablation}. We focus on MF-en, a retrieval-intensive task in LongBench, because it requires long-context models to identify a small amount of useful evidence from a large prefix. This makes MF-en a direct probe for both inter-chunk sparsity and intra-chunk sparsity.

\paragraph{Comparison with BM25 retrieval.}
Table~\ref{tab:fair_budget} compares our predictive selection with BM25 context selection at equal selected-token budgets, using the corresponding base model for generation. \ours{} outperforms BM25 in several settings, particularly with intra-chunk sparsity on Dream-7B and inter-chunk selection on UltraLLaDA. BM25 remains competitive in other settings, so these results do not imply that predictive selection uniformly dominates lexical retrieval.

\begin{table}[!ht]
\centering
\scriptsize
\setlength{\tabcolsep}{3pt}
\renewcommand{\arraystretch}{1.1}
\caption{MF-en performance under matched selected-token budgets on the full 150-example test set. Higher is better. For each base model, BM25 is used only for context selection, while generation is performed by the same Dream-7B or UltraLLaDA model used by our method.}
\label{tab:fair_budget}
\begin{tabular}{llrr}
\toprule
Base model & Selection & Budget & MF-en \\
\midrule
\multirow{4}{*}{Dream-7B}
& BM25 & 2K & 41.28 \\
& BM25 & 4K & 38.41 \\
& Ours (inter) & 4K & 38.26 \\
& Ours (inter + intra) & 2K & \textbf{46.40} \\
\midrule
\multirow{4}{*}{UltraLLaDA}
& BM25 & 2K & 33.96 \\
& BM25 & 4K & 35.13 \\
& Ours (inter) & 2K & \textbf{39.94} \\
& Ours (inter + intra) & 4K & \textbf{36.07} \\
\bottomrule
\end{tabular}
\end{table}

\paragraph{Multi-hop KV cache compression evaluation.}
On SCBench (Table~\ref{tab:scbench}), \ours{} achieves the best Overall Accuracy and Passkey Accuracy, while BM25 + Fast-dLLM v1 achieves higher Overall F1 and Summary F1. This gap suggests that static selection may be less effective for aggregation-heavy summarization than for retaining a small set of exact evidence.

\begin{table}[!ht]
\centering
\scriptsize
\setlength{\tabcolsep}{2.8pt}
\renewcommand{\arraystretch}{1.1}
\caption{Performance on the SCBench Summary-with-Needles task, which jointly evaluates summarization and needle retrieval. All methods retain 8K selected context tokens for decoding. Higher values are better.}
\label{tab:scbench}
\resizebox{\columnwidth}{!}{%
\begin{tabular}{lcccc}
\toprule
Method & Overall Acc. & Overall F1 & Passkey Acc. & Summary F1 \\
\midrule
BM25 + Vanilla & 0.2875 & 0.4490 & 0.5750 & 0.3230 \\
BM25 + Fast-dLLM v1 & 0.3125 & \textbf{0.5074} & 0.6250 & \textbf{0.3898} \\
BM25 + dKV-Cache & 0.3000 & 0.4746 & 0.6000 & 0.3492 \\
BM25 + Sparse-dLLM & 0.2875 & 0.4477 & 0.5750 & 0.3204 \\
\rowcolor{mycolor!30}
Ours & \textbf{0.3250} & 0.4996 & \textbf{0.6500} & 0.3492 \\
\bottomrule
\end{tabular}%
}
\end{table}

\paragraph{Ablation on top-$K$.}
The optimal number of selected chunks is small, but it depends on whether intra-chunk sparsity is used. Without intra-chunk sparsity, Dream-v0-Base-7B performs best with top-$K=4$, while UltraLLaDA prefers top-$K=2$ on the $o46$ subset. With intra-chunk sparsity, Dream still favors top-$K=4$, whereas UltraLLaDA improves with top-$K=8$, suggesting that token-level compression can benefit from a broader candidate chunk pool.

\paragraph{Ablation on chunk size.}
The best chunk granularity is model-dependent. Dream-v0-Base-7B favors $C=1024$, which balances retrieval resolution with enough local context inside each selected chunk. UltraLLaDA, evaluated on the $o46$ subset, performs best at $C=2048$ in the top-$K=4$ sweep, indicating that the native 128K model can benefit from slightly coarser chunks in this setting.

\paragraph{Ablation on chunk BOS.}
The chunk-BOS control shows that boundary handling matters, but its effect is not uniform. On UltraLLaDA, keeping the chunk BOS improves MF-en from 28.06 to 29.34 under the top-$K=4$, $C=1024$ self-information setting. On Dream-v0-Base-7B, the effect is mixed in the early \texttt{chunk-query} route: removing the chunk BOS helps with 16 pseudo-labels, while keeping it is better with 4 pseudo-labels. We therefore keep chunk BOS in the final configuration.

\paragraph{Ablation on chunk score.}
Pseudo-label scoring is a stable way to rank chunks before final decoding. On Dream-v0-Base-7B, short pseudo-label windows slightly improve over self-information, while longer pseudo-label windows do not consistently help. On UltraLLaDA, one or two pseudo-labels already improve over self-information on the $o46$ subset, and increasing the pseudo-label window brings no additional gain.

\paragraph{Ablation on pseudo-label denoising steps.}
Pseudo-label denoising should use only a few steps. Dream-v0-Base-7B benefits from two denoising steps in the tested pseudo-label settings, but additional steps reduce the score. UltraLLaDA shows an even stronger preference for early pseudo-labels: one denoising step is best across the tested pseudo-label windows, and further steps degrade performance, suggesting that repeated refinement can inject noise into the chunk-ranking signal.

\paragraph{Ablation on KV cache construction strategy.}
We compare three strategies for constructing the KV cache after chunk selection. In \texttt{chunk-only}, each selected chunk is encoded independently without the query. In \texttt{chunk-query}, each selected chunk is encoded together with the query, while different chunks remain isolated from one another. In \texttt{full-mask}, all selected chunks and the query are jointly encoded using the same bidirectional attention mask as in the decoding stage. On Dream-v0-Base-7B, \texttt{full-mask} substantially outperforms \texttt{chunk-query} and \texttt{chunk-only}, showing that aligning cache construction with the decoding-time attention pattern is important.

\paragraph{Ablation on retained-token budget.}
When intra-chunk sparsity is enabled, both base models benefit from a larger retained-token budget. Increasing the budget from $B=256$ to $B=512$ improves Dream-v0-Base-7B and UltraLLaDA, supporting our choice of a moderate top-$B$ budget that preserves useful local evidence inside each selected chunk.

\paragraph{Ablation on token-retention score.}
For Dream-v0-Base-7B, the bidirectional token-retention score improves over the query-to-chunk score, supporting our design choice of measuring token importance using attention in both directions between query tokens and chunk tokens. Overall, the ablations validate the main configuration used in our experiments: inter-chunk sparsity selects a compact set of relevant chunks, while intra-chunk sparsity preserves the most query-relevant tokens with a fixed top-$B$ budget.

\begin{table*}[p]
\centering
\scriptsize
\setlength{\tabcolsep}{2pt}
\renewcommand{\arraystretch}{1.15}
\caption{Ablation results for Dream-v0-Base-7B on the LongBench MF-en task without intra-chunk sparsity (full test set, $n=150$). Each block varies one design choice; shaded cells indicate the varied parameter.}
\label{tab:dream_mfen_ablation}
\begin{tabular}{lrrclrllr}
\toprule
Variant & top-$K$ & $C$ & chunk BOS & chunk score & \# pseudo-labels & denoising steps & KV build & MF-en \\
\midrule
\multicolumn{9}{l}{\textit{Ablation on top-$K$}} \\
\midrule
top-$K=4$ & \cellcolor{mycolor!30}4 & 1024 & on & \texttt{pseudo-label} & 4 & -- & \texttt{full-mask} & 46.57 \\
top-$K=6$ & \cellcolor{mycolor!30}6 & 1024 & on & \texttt{pseudo-label} & 4 & -- & \texttt{full-mask} & 45.01 \\
\midrule
\multicolumn{9}{l}{\textit{Ablation on chunk size}} \\
\midrule
chunk 1024 & 4 & \cellcolor{mycolor!30}1024 & on & \texttt{pseudo-label} & 4 & -- & \texttt{full-mask} & 41.52 \\
chunk 1900 & 4 & \cellcolor{mycolor!30}1900 & on & \texttt{pseudo-label} & 4 & -- & \texttt{full-mask} & 37.68 \\
chunk 512 & 4 & \cellcolor{mycolor!30}512 & on & \texttt{pseudo-label} & 4 & -- & \texttt{full-mask} & 33.14 \\
\midrule
\multicolumn{9}{l}{\textit{Ablation on chunk BOS (chunk-query KV)}} \\
\midrule
draft16 BOS on & 4 & 1024 & \cellcolor{mycolor!30}on & \texttt{pseudo-label} & 16 & -- & \texttt{chunk-query} & 39.32 \\
draft16 BOS off & 4 & 1024 & \cellcolor{mycolor!30}off & \texttt{pseudo-label} & 16 & -- & \texttt{chunk-query} & 40.81 \\
draft4 BOS on & 4 & 1024 & \cellcolor{mycolor!30}on & \texttt{pseudo-label} & 4 & -- & \texttt{chunk-query} & 43.39 \\
draft4 BOS off & 4 & 1024 & \cellcolor{mycolor!30}off & \texttt{pseudo-label} & 4 & -- & \texttt{chunk-query} & 42.73 \\
\midrule
\multicolumn{9}{l}{\textit{Ablation on chunk score}} \\
\midrule
query-only & 4 & 1024 & on & \cellcolor{mycolor!30}\texttt{self-info} & \cellcolor{mycolor!30}0 & -- & \texttt{full-mask} & 46.65 \\
pseudo-2 & 4 & 1024 & on & \cellcolor{mycolor!30}\texttt{pseudo-label} & \cellcolor{mycolor!30}2 & 1 & \texttt{full-mask} & 46.89 \\
pseudo-4 & 4 & 1024 & on & \cellcolor{mycolor!30}\texttt{pseudo-label} & \cellcolor{mycolor!30}4 & 1 & \texttt{full-mask} & 46.57 \\
pseudo-8 & 4 & 1024 & on & \cellcolor{mycolor!30}\texttt{pseudo-label} & \cellcolor{mycolor!30}8 & 1 & \texttt{full-mask} & 46.15 \\
pseudo-16 & 4 & 1024 & on & \cellcolor{mycolor!30}\texttt{pseudo-label} & \cellcolor{mycolor!30}16 & 1 & \texttt{full-mask} & 45.99 \\
pseudo-32 & 4 & 1024 & on & \cellcolor{mycolor!30}\texttt{pseudo-label} & \cellcolor{mycolor!30}32 & 1 & \texttt{full-mask} & 46.68 \\
\midrule
\multicolumn{9}{l}{\textit{Ablation on pseudo-label denoising steps (2 pseudo-labels)}} \\
\midrule
1 step & 4 & 1024 & on & \texttt{pseudo-label} & 2 & \cellcolor{mycolor!30}1 & \texttt{full-mask} & 46.48 \\
2 steps & 4 & 1024 & on & \texttt{pseudo-label} & 2 & \cellcolor{mycolor!30}2 & \texttt{full-mask} & 46.89 \\
\midrule
\multicolumn{9}{l}{\textit{Ablation on pseudo-label denoising steps (4 pseudo-labels)}} \\
\midrule
2 steps & 4 & 1024 & on & \texttt{pseudo-label} & 4 & \cellcolor{mycolor!30}2 & \texttt{full-mask} & \textbf{47.54} \\
3 steps & 4 & 1024 & on & \texttt{pseudo-label} & 4 & \cellcolor{mycolor!30}3 & \texttt{full-mask} & 46.41 \\
4 steps & 4 & 1024 & on & \texttt{pseudo-label} & 4 & \cellcolor{mycolor!30}4 & \texttt{full-mask} & 46.57 \\
\midrule
\multicolumn{9}{l}{\textit{Ablation on KV cache construction strategy}} \\
\midrule
chunk-query KV & 4 & 1024 & on & \texttt{pseudo-label} & 4 & 2 & \cellcolor{mycolor!30}\texttt{chunk-query} & 41.52 \\
chunk-only KV & 4 & 1024 & on & \texttt{pseudo-label} & 4 & 2 & \cellcolor{mycolor!30}\texttt{chunk-only} & 34.13 \\
full-mask KV & 4 & 1024 & on & \texttt{pseudo-label} & 4 & 2 & \cellcolor{mycolor!30}\texttt{full-mask} & 46.57 \\
\bottomrule
\end{tabular}
\end{table*}

\begin{table*}[p]
\centering
\scriptsize
\setlength{\tabcolsep}{2pt}
\renewcommand{\arraystretch}{1.15}
\caption{Ablation results for Dream-v0-Base-7B on the LongBench MF-en task with intra-chunk sparsity (full test set, $n=150$). Each block varies one design choice; shaded cells indicate the varied parameter.}
\label{tab:dream_mfen_eviction_ablation}
\begin{tabular}{lrrlrrlrlr}
\toprule
Variant & top-$K$ & $C$ & chunk score & \# pseudo-labels & denoising steps & KV build & top-$B$ & token score & MF-en \\
\midrule
\multicolumn{10}{l}{\textit{Ablation on top-$K$}} \\
\midrule
top-$K=6$ & \cellcolor{mycolor!30}6 & 1024 & \texttt{pseudo-label} & 4 & 2 & \texttt{full-mask} & 512 & \texttt{query-to-chunk} & 45.21 \\
top-$K=4$ & \cellcolor{mycolor!30}4 & 1024 & \texttt{pseudo-label} & 4 & 2 & \texttt{full-mask} & 512 & \texttt{query-to-chunk} & 47.54 \\
top-$K=2$ & \cellcolor{mycolor!30}2 & 1024 & \texttt{pseudo-label} & 4 & 2 & \texttt{full-mask} & 512 & \texttt{query-to-chunk} & 46.94 \\
\midrule
\multicolumn{10}{l}{\textit{Ablation on retained-token budget}} \\
\midrule
cap 256 & 4 & 1024 & \texttt{pseudo-label} & 4 & 2 & \texttt{full-mask} & \cellcolor{mycolor!30}256 & \texttt{query-to-chunk} & 44.46 \\
cap 512 & 4 & 1024 & \texttt{pseudo-label} & 4 & 2 & \texttt{full-mask} & \cellcolor{mycolor!30}512 & \texttt{query-to-chunk} & 47.54 \\
\midrule
\multicolumn{10}{l}{\textit{Ablation on token-retention score}} \\
\midrule
query-to-chunk & 4 & 1024 & \texttt{pseudo-label} & 4 & 2 & \texttt{full-mask} & 512 & \cellcolor{mycolor!30}\texttt{query-to-chunk} & 47.54 \\
bidirectional & 4 & 1024 & \texttt{pseudo-label} & 4 & 2 & \texttt{full-mask} & 512 & \cellcolor{mycolor!30}\texttt{bidirectional} & \textbf{47.96} \\
\bottomrule
\end{tabular}
\end{table*}

\begin{table*}[p]
\centering
\scriptsize
\setlength{\tabcolsep}{2pt}
\renewcommand{\arraystretch}{1.15}
\caption{Ablation results for UltraLLaDA on the LongBench MF-en task without intra-chunk sparsity ($o46$ subset, $n=46$). Each block varies one design choice; shaded cells indicate the varied parameter.}
\label{tab:ultrallada_mfen_ablation}
\begin{tabular}{lrrclrllr}
\toprule
Variant & top-$K$ & $C$ & chunk BOS & chunk score & \# pseudo-labels & denoising steps & KV build & MF-en o46 \\
\midrule
\multicolumn{9}{l}{\textit{Ablation on top-$K$ ($C=1024$)}} \\
\midrule
top-$K=2$ & \cellcolor{mycolor!30}2 & 1024 & on & \texttt{self-info} & 0 & -- & \texttt{full-mask} & 34.84 \\
top-$K=3$ & \cellcolor{mycolor!30}3 & 1024 & on & \texttt{self-info} & 0 & -- & \texttt{full-mask} & 29.08 \\
top-$K=4$ & \cellcolor{mycolor!30}4 & 1024 & on & \texttt{self-info} & 0 & -- & \texttt{full-mask} & 29.34 \\
top-$K=5$ & \cellcolor{mycolor!30}5 & 1024 & on & \texttt{self-info} & 0 & -- & \texttt{full-mask} & 29.19 \\
top-$K=6$ & \cellcolor{mycolor!30}6 & 1024 & on & \texttt{self-info} & 0 & -- & \texttt{full-mask} & 28.66 \\
top-$K=8$ & \cellcolor{mycolor!30}8 & 1024 & on & \texttt{self-info} & 0 & -- & \texttt{full-mask} & 26.29 \\
\midrule
\multicolumn{9}{l}{\textit{Ablation on chunk size (top-$K=4$)}} \\
\midrule
chunk 512 & 4 & \cellcolor{mycolor!30}512 & on & \texttt{self-info} & 0 & -- & \texttt{full-mask} & 26.62 \\
chunk 1024 & 4 & \cellcolor{mycolor!30}1024 & on & \texttt{self-info} & 0 & -- & \texttt{full-mask} & 29.34 \\
chunk 2048 & 4 & \cellcolor{mycolor!30}2048 & on & \texttt{self-info} & 0 & -- & \texttt{full-mask} & 31.51 \\
chunk 4096 & 4 & \cellcolor{mycolor!30}4096 & on & \texttt{self-info} & 0 & -- & \texttt{full-mask} & 27.34 \\
\midrule
\multicolumn{9}{l}{\textit{Ablation on chunk BOS}} \\
\midrule
chunk BOS on & 4 & 1024 & \cellcolor{mycolor!30}on & \texttt{self-info} & 0 & -- & \texttt{full-mask} & 29.34 \\
chunk BOS off & 4 & 1024 & \cellcolor{mycolor!30}off & \texttt{self-info} & 0 & -- & \texttt{full-mask} & 28.06 \\
\midrule
\multicolumn{9}{l}{\textit{Ablation on chunk score}} \\
\midrule
query-only & 2 & 1024 & on & \cellcolor{mycolor!30}\texttt{self-info} & \cellcolor{mycolor!30}0 & -- & \texttt{full-mask} & 34.84 \\
pseudo-1 & 2 & 1024 & on & \cellcolor{mycolor!30}\texttt{pseudo-label} & \cellcolor{mycolor!30}1 & 1 & \texttt{full-mask} & \textbf{37.22} \\
pseudo-2 & 2 & 1024 & on & \cellcolor{mycolor!30}\texttt{pseudo-label} & \cellcolor{mycolor!30}2 & 1 & \texttt{full-mask} & 37.17 \\
pseudo-4 & 2 & 1024 & on & \cellcolor{mycolor!30}\texttt{pseudo-label} & \cellcolor{mycolor!30}4 & 1 & \texttt{full-mask} & 36.17 \\
pseudo-8 & 2 & 1024 & on & \cellcolor{mycolor!30}\texttt{pseudo-label} & \cellcolor{mycolor!30}8 & 1 & \texttt{full-mask} & 36.39 \\
\midrule
\multicolumn{9}{l}{\textit{Ablation on pseudo-label denoising steps (2 pseudo-labels)}} \\
\midrule
1 step & 2 & 1024 & on & \texttt{pseudo-label} & 2 & \cellcolor{mycolor!30}1 & \texttt{full-mask} & 37.17 \\
2 steps & 2 & 1024 & on & \texttt{pseudo-label} & 2 & \cellcolor{mycolor!30}2 & \texttt{full-mask} & 34.75 \\
\midrule
\multicolumn{9}{l}{\textit{Ablation on pseudo-label denoising steps (4 pseudo-labels)}} \\
\midrule
1 step & 2 & 1024 & on & \texttt{pseudo-label} & 4 & \cellcolor{mycolor!30}1 & \texttt{full-mask} & 36.17 \\
2 steps & 2 & 1024 & on & \texttt{pseudo-label} & 4 & \cellcolor{mycolor!30}2 & \texttt{full-mask} & 33.53 \\
3 steps & 2 & 1024 & on & \texttt{pseudo-label} & 4 & \cellcolor{mycolor!30}3 & \texttt{full-mask} & 31.53 \\
4 steps & 2 & 1024 & on & \texttt{pseudo-label} & 4 & \cellcolor{mycolor!30}4 & \texttt{full-mask} & 33.07 \\
\midrule
\multicolumn{9}{l}{\textit{Ablation on pseudo-label denoising steps (8 pseudo-labels)}} \\
\midrule
1 step & 2 & 1024 & on & \texttt{pseudo-label} & 8 & \cellcolor{mycolor!30}1 & \texttt{full-mask} & 36.39 \\
2 steps & 2 & 1024 & on & \texttt{pseudo-label} & 8 & \cellcolor{mycolor!30}2 & \texttt{full-mask} & 32.35 \\
\bottomrule
\end{tabular}
\end{table*}

\begin{table*}[p]
\centering
\scriptsize
\setlength{\tabcolsep}{2pt}
\renewcommand{\arraystretch}{1.15}
\caption{Ablation results for UltraLLaDA on the LongBench MF-en task with intra-chunk sparsity ($o46$ subset, $n=46$). Each block varies one design choice; shaded cells indicate the varied parameter.}
\label{tab:ultrallada_mfen_eviction_ablation}
\begin{tabular}{lrrlrrlrlr}
\toprule
Variant & top-$K$ & $C$ & chunk score & \# pseudo-labels & denoising steps & KV build & top-$B$ & token score & MF-en o46 \\
\midrule
\multicolumn{10}{l}{\textit{Ablation on retained-token budget}} \\
\midrule
cap 256 & 8 & 1024 & \texttt{self-info} & 0 & -- & \texttt{full-mask} & \cellcolor{mycolor!30}256 & \texttt{bidirectional} & 26.01 \\
cap 512 & 8 & 1024 & \texttt{self-info} & 0 & -- & \texttt{full-mask} & \cellcolor{mycolor!30}512 & \texttt{bidirectional} & \textbf{28.75} \\
\midrule
\multicolumn{10}{l}{\textit{Ablation on top-$K$ with intra-chunk sparsity}} \\
\midrule
top-$K=4$ & \cellcolor{mycolor!30}4 & 1024 & \texttt{self-info} & 0 & -- & \texttt{full-mask} & 512 & \texttt{bidirectional} & 27.71 \\
top-$K=6$ & \cellcolor{mycolor!30}6 & 1024 & \texttt{self-info} & 0 & -- & \texttt{full-mask} & 512 & \texttt{bidirectional} & 27.62 \\
top-$K=8$ & \cellcolor{mycolor!30}8 & 1024 & \texttt{self-info} & 0 & -- & \texttt{full-mask} & 512 & \texttt{bidirectional} & \textbf{28.75} \\
\bottomrule
\end{tabular}
\end{table*}

\section{Prefilling Efficiency Analysis}
\label{sec:appendix_efficiency}

\begin{figure}[t]
    \centering
    \includegraphics[width=0.85\columnwidth]{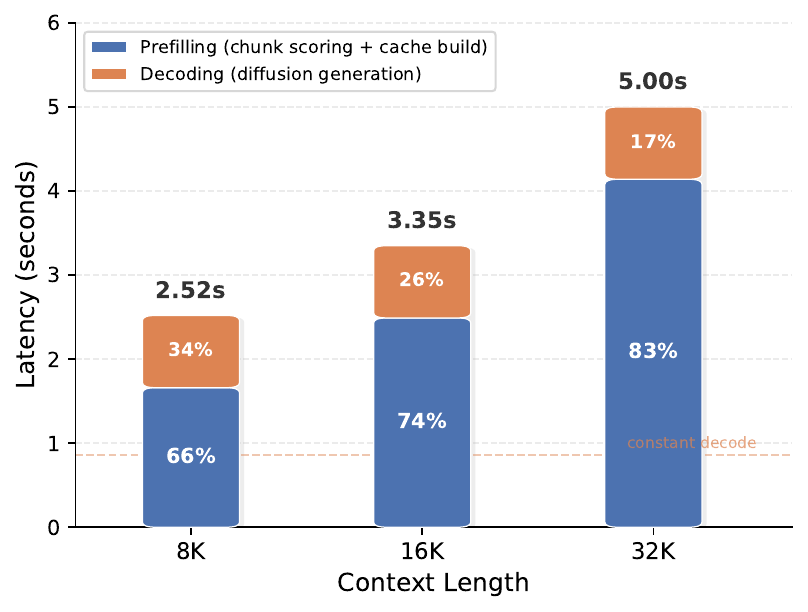}
    \caption{Latency breakdown of \ours{} into prefilling (chunk scoring + cache build) and decoding (diffusion generation). Decoding time remains constant ($\sim$0.86s) regardless of context length, while prefilling scales with input size.}
    \label{fig:breakdown}
\end{figure}

\paragraph{Efficiency sensitivity.}
Table~\ref{tab:efficiency_sensitivity} shows that speedup increases with output length because the fixed scoring overhead is amortized over more denoising work. Larger batches reduce the relative speedup, but \ours{} remains substantially faster across the tested batch sizes.

\begin{table}[!ht]
\centering
\scriptsize
\setlength{\tabcolsep}{4pt}
\renewcommand{\arraystretch}{1.1}
\caption{End-to-end speedup at 32K context under varying output lengths and batch sizes. Input prefix and scoring configuration are fixed.}
\label{tab:efficiency_sensitivity}
\begin{tabular}{lrrrr}
\toprule
Output length & 32 & 64 & 128 & 256 \\
\midrule
Speedup ($\times$) & 28.0 & 33.5 & 38.7 & 42.1 \\
\bottomrule
\end{tabular}
\vspace{1.5mm}
\begin{tabular}{lrrrr}
\toprule
Batch size & 1 & 2 & 4 & 8 \\
\midrule
Speedup ($\times$) & 28.0 & 28.6 & 22.8 & 19.4 \\
\bottomrule
\end{tabular}
\end{table}

To isolate the contribution of each phase, we separately measure the latency of \emph{prefilling} (chunk scoring + KV cache construction) and \emph{decoding} (diffusion generation) within \ours{}. As shown in Figure~\ref{fig:breakdown}, the decoding latency remains nearly constant ($\sim$0.86s) across all context lengths, since it always operates on the fixed compressed context ($\sim$4K tokens). The prefilling cost grows with input length (1.66s at 8K, 2.49s at 16K, 4.14s at 32K) as chunk scoring must attend over the full context. Nevertheless, prefilling is a one-time cost amortized over the entire generation, and the constant decoding time explains why \ours{}'s speedup advantage widens at longer contexts.

\end{document}